\pdfoutput=1
\documentclass[journal]{IEEEtran}
\usepackage[OT1]{fontenc}

\ifCLASSOPTIONcompsoc
\usepackage[caption=false, font=normalsize, labelfont=sf, textfont=sf]{subfig}
\else
\usepackage[caption=false, font=footnotesize]{subfig}
\fi
\usepackage[export]{adjustbox}
\usepackage{float}
\usepackage{graphicx}
\usepackage{booktabs}
\usepackage{multirow}
\usepackage{threeparttable}
\usepackage{etoolbox}
\usepackage{multicol}
\usepackage[margin=2cm]{geometry}
\usepackage{lipsum}
\usepackage{microtype}
\usepackage{vwcol} 
\usepackage{amsmath}
\usepackage{algorithmicx}
\usepackage{algorithm}
\usepackage{algpseudocode}
\usepackage{amssymb}
\usepackage{multicol}
\usepackage{array}
\usepackage[switch]{lineno}

\begin{document}

\title{Scene-Aware Error Modeling of LiDAR/Visual Odometry for Fusion-based Vehicle Localization}
\author{Xiaoliang Ju$^{1}$, Donghao Xu$^{1}$, Huijing Zhao$^{1}$
	\thanks{This work is partially supported by the NSFC Grants 61973004.}
	\thanks{$^{1}$X. Ju, D. Xu, and H. Zhao are with the Peking University, with the
		Key Laboratory of Machine Perception (MOE), and also with the School
		of Electronics Engineering and Computer Science, Beijing 100871, China
}
\thanks{Correspondence: H. Zhao, zhaohj@pku.edu.cn.}}

\maketitle

\begin{abstract}
Localization is an essential technique in mobile robotics.
In a complex environment, it is necessary to fuse different localization modules to obtain more robust results, in which the error model plays a paramount role.
However, exteroceptive sensor-based odometries (ESOs), such as LiDAR/visual odometry, often deliver results with scene-related error, which is difficult to model accurately.
To address this problem, this research designs a scene-aware error model for ESO, based on which a multimodal localization fusion framework is developed. In addition, an end-to-end learning method is proposed to train this error model using sparse global poses such as GPS/IMU results.
The proposed method is realized for error modeling of LiDAR/visual odometry, and the results are fused with dead reckoning to examine the performance of vehicle localization. Experiments are conducted using both simulation and real-world data of experienced and unexperienced environments, and the experimental results demonstrate that with the learned scene-aware error models, vehicle localization accuracy can be largely improved and shows adaptiveness in unexperienced scenes.
\end{abstract}


\IEEEpeerreviewmaketitle
\section{Introduction}
Recent years have witnessed considerable progress in developing autonomous systems\cite{hebert2012intelligent}\cite{jones2006robots}\cite{thrun2006stanley}\cite{nelson2019petman}, where highly accurate vehicle localization is the key to achieving safe and efficient autonomy in a complex real world.

GNSSs (global navigation satellite systems) have been widely used in vehicle localization in outdoor environments and are usually combined with proprioceptive sensors such as IMUs (inertial measurement units) and wheel encoders for interpolating positions during satellite signal outages\cite{georgy2011enhanced}\cite{ndjeng2011low}. However, such systems are restricted by GNSS conditions, and the IMU maintains accuracy only for short periods due to accelerometer biases and gyro drifts. Therefore, exteroceptive sensor-based approaches such as LiDAR odometry\cite{biber2003normal}\cite{Olson2009}\cite{Zhang2014} or visual odometry\cite{Kitt2010}\cite{Forster2014}\cite{Engel2017} have been studied to assist in highly accurate localization. Hereinafter, we refer to exteroceptive sensor-based odometry as ESO and proprioceptive sensor-based odometry as PSO.

However, the performance of exteroceptive sensor-based localization is strongly related to the scenes. When fusing them with other localization approaches, e.g., \cite{Chiu2014}\cite{Sukumar2007}\cite{Bresson2016}, precise error modeling is essential to the fusion efficiency. Covariance has been a widely used measurement for error estimation\cite{Thrun2005}\cite{Shen2014}\cite{Koch2016}. Many of these methods correlate pose uncertainty with the covariance of data matching\cite{Bengtsson2003}\cite{Bonnabel2016}, which could be an ill-posed problem in many situations. In addition, most existing works only model the error of their measurements or features\cite{Bloesch2015}\cite{Hossein2010}\cite{Gulati2016}, whereas fewer studies focus on the error of final localization results such as \cite{Hemann2016}\cite{Koch2016}\cite{Lynen2013}.

This research proposes a method of ESO scene-aware error modeling for fusion-based localization,
which is formulated as a mapping from given scene data to a prediction of the ESO error as an information matrix, a dual form of a covariance matrix.
A CNN (convolutional neural network) is used to model the mapping procedure, and a vehicle localization framework is devised to incorporate the scene-aware error modeling results in fusing ESO for pose estimation.
An end-to-end method is developed to train the error model by using reliable global localization, such as GPS, as supervision, which could be sporadic. Therefore, at each iteration, the vehicle is localized by forward propagation on the current parameters for a number of frames, and when a reliable GPS measurement is found, the global localization error is backpropagated along the pipeline to correct the CNN parameters.

The proposed method is realized for error modeling of LiDAR odometry and visual odometry, and the results are fused with dead reckoning to examine the performance of vehicle localization.
Experiments are conducted using both simulation and real-world data. The former validates the adaptability of the proposed method in simple but typical scenes, while the latter examines the performance in a complex real world that contains experienced and unexperienced environments. The experiments are deployed on some popular LiDAR odometry\cite{Olson2009}\cite{Censi2008} and visual odometry\cite{Kitt2010}\cite{Engel2017}\cite{Gomez-ojeda2016} methods and compared with the traditional fusion approaches using covariance-based error modeling. The experimental results demonstrate that with the learned scene-aware error models, vehicle localization accuracy can be largely improved, and it shows adaptiveness in unexperienced scenes.

The paper is organized as follows. A literature review about ESO and the corresponding error model is presented in section \ref{relatedworks}. An overview of the proposed error model learning method is given in section \ref{method}. Experiments on LiDAR/visual odometry and the analysis of their results are illustrated in sections \ref{experiment} and \ref{sec:VOExp}, respectively.
Finally, section
\ref{conclusion} describes the conclusions and the direction of our future works.

\begin{table*}[htbp]
\centering
\resizebox{1.0\textwidth}{!}{%
\begin{threeparttable}[b]
\caption{Exteroceptive Sensor Based Odometry}
\label{table:PBO}
\def\arraystretch{1.3}
\begin{tabular}{lllllll}
\hline
 & Research & Category & Optimizing Method & Feature & Objective & Error Model \\ \hline
\multirow{13}{*}{LO} 
& Censi, 2008, \cite{Censi2008} & F-b & Lagrange's Multiplier & RP\&Line & Feature Distance &  \\
 & Bosse, 2009, \cite{Bosse2009} & F-b & WLS & Shape Info. & Match\&Smoothness Func. & - \\
 & Armesto, 2010, \cite{Armesto2010} & F-b & LS & RP, Facet & Metric Distance &  \\
 & Zhang, 2014, \cite{Zhang2014} & F-b & LM & Line, Plane & Feature Distance & - \\
 & Velas, 2016, \cite{Velas2016} & F-b & SVD & Line & Feature Distance & - \\
 & Wang, 2018, \cite{Wang2018} & F-b & EM, IRLS & RP & Likelihood Func. & - \\
 & Magnusson, 2007, \cite{Magnusson2007} & Direct & Newton & RP & Joint Prob. & - \\
 & Olson, 2009, \cite{Olson2009} & Direct & Search & RP & Posterior Observation Prob. & Cov. \\
 & Olson, 2015, \cite{Olson2015} & Direct & Search & RP & Correlative Cost Func. & - \\
 & Jaimez, 2016, \cite{Jaimez2016} & Direct & IRLS & RP(Range Flow) & Geometric Residual & - \\
 & Ramos, 2007, \cite{Ramos2007} & Heuristic/F-b & WLS & Local Feature & CRF Inference Error & - \\
 & Diosi, 2007, \cite{Diosi2007} & Heuristic/F-b & WLS, Parabola Fitting & RP & Polar Range Distance & Cov. \\
 & Censi, 2009, \cite{Censi2009} & Heuristic/Direct & LS & RP(Hough) & Spectrum Func. & - \\ \hline
\multirow{12}{*}{VO} 
 & Howard, 2008, \cite{Howard2008} & F-b & LM & Harris/FAST & RE & - \\
 & Kitt, 2010, \cite{Kitt2010} & F-b & ISPKF & Harris et.al. & RE & Cov. \\
 & Mouats, 2014, \cite{Mouats2014} & F-b & GN & Log-Gabor Wavelets & RE & - \\
 & Gomez-ojeda, 2016, \cite{Gomez-ojeda2016} & F-b & GN & ORB\&LSD & RE & Cov. \\

 & Zhang, 2012, \cite{Zhang2012} & F-b & PHD & SIFT & RE & Cov. \\
 & Engel, 2013, \cite{Engel2013} & Direct & RGN & - & PE & Cov. \\
 & Kerl, 2013, \cite{Kerl2013} & Direct & IRLS & - & PE & - \\
 & Wang, 2017, \cite{Wang2017} & Direct & GN & - & PE & - \\
 & Li, 2018, \cite{Li2018} & Direct & GN & - & PE & - \\
  & Engel, 2017, \cite{Engel2017} & Direct & GN & - & PE & - \\
 & Forster, 2014, \cite{Forster2014} & Semi-Direct & GN & Sparse Feature Patches & PE\&RE & - \\
  & Wang, 2017, \cite{Wang2017deep} & Deep Learning& BP & - & Pose MSE & - \\

 \hline
\multirow{6}{*}{Others} 

 & Tanskanen, 2015, \cite{Tanskanen2015} & Visual-Inertial & EKF & - & PE & Cov. \\
 & Usenko, 2016, \cite{Usenko2016} & Visual-Inertial & LM & - & Photometric-Inertial Energy & - \\
  & Qin, 2018, \cite{Qin2018} & Visual-Inertial & LS & Harris & Feature\&IMU Residual & - \\
   & Zhang, 2015, \cite{Zhang2015} & Visual-LiDAR & LM & Haris\&RP & Feature Distance & - \\
 & Hemann, 2016, \cite{Hemann2016} & LiDAR-Inertial & KF & RP\&DEM & Cross-correlation Func. & Cov. \\
 & Barjenbruch, 2015, \cite{Barjenbruch2015} & Radar & Gradient-based  & Spatial\&Doppler Info. & Metric Func. & - \\

 \bottomrule
\end{tabular}%
\begin{tablenotes}
	\item *The denotions for abbreviations in this table are arranged in alphabetical order by column. 
\end{tablenotes}
\setlength{\columnsep}{0.9cm}
    \setlength{\multicolsep}{0.0cm}
   \begin{multicols}{4}
   \begin{tablenotes}
   \scriptsize
	\item BP: Back Propagation
	\item Cov.: Covariance
	\item CRF: Conditional Random Field
	\item DEM: Digital Elevation Model
	\item EKF: Extended Kalman Filter
	\item EM: Expectation-Maximization 
	\item F-b: Feature-based
	\item Func.: Function
	\item GN: Gauss-Newton 
	\item IMU: Inertial Measurement Unit
	\item Info.: Information
	\item IRLS: Iteratively Reweighted Least Squares  
	\item ISPKF: Iterated Sigma Point Kalman Filter
	\item KF: Kalman Filter
	\item LM: Levenberg-Marquardt 
	\item LO: LiDAR Odometry 
	\item LS: Least Squares 
	\item MSE: Mean-square Error
	\item PE: Photometric Error
	\item PHD: Probability Hypothesis Density
	\item Prob.: Probability
	\item RE: Reprojection Error
	\item RGN: Reweighted Gauss-Newton
	\item RP: Raw Point
	\item SVD: Singular Value Decomposition 
	\item VO: Visual Odometry 
	\item WLS: Weighted Least Squares 
   \end{tablenotes}
     \end{multicols}
\end{threeparttable}

}%resizebox
\end{table*}

\section{Related Works}
\label{relatedworks}

\subsection{LiDAR Odometry}
LiDAR odometry performs relative positioning by comparing laser measurements from sequent LiDAR scans, which has a more popular name, scan matching. Following the conventional taxonomy of visual odometry, this paper divides LiDAR odometries into feature-based methods and direct methods by whether explicit feature correspondence is needed.

\textbf{Feature-based methods.} A typical method for scan matching is to build the feature correspondence for sequent LiDAR scans, and then the motion from the reference frame to the target frame can be calculated from the matching results.
In feature selection, various definitions of features, such as points, lines, planes and other self-defined local features, can be used alone or in combination\cite{Censi2008}\cite{Zhang2014}.
In optimization strategies of feature matching, many works, such as \cite{Bosse2009}and \cite{Armesto2010}, are variants of the ICP (iterative closest point)\cite{besl1992method} algorithm, which iteratively minimizes the feature matching error using an optimizer such as least squares. Apparently, feature association in such an indirect matching method creates considerable computing cost and often leads to overconfident mismatching.

\textbf{Direct methods.} To overcome the efficiency problem of feature association, some researchers have attempted to avoid building such explicit correspondence. \cite{biber2003normal} transformed the scan-to-scan matching problem into a correlation evaluation under a probabilistic framework, and \cite{Magnusson2007} extended it to 3D applications. \cite{Olson2009} proposed correlative scan matching by employing a Monte Carlo sampling strategy, and \cite{Olson2015} improved the efficiency of such methods using multiresolution matching. \cite{Jaimez2016} designed a range flow-based approach in the fashion of dense 3D visual odometry, which performs scan alignment using scan gradients.

In addition, many heuristic methods have been proposed to compensate for the flaws of previous work, such as poor convergence or dependence on initialization. \cite{Diosi2007} matched LiDAR points with the same bearing under polar coordinates to run faster than ICP. \cite{Ramos2007} presented a CRF (conditional random field)-based scan matching, which takes into account the high-level shape information. \cite{Censi2009} attempted to use the Hough transformation to decompose the 6DoF search into a series of fast one-dimensional cross-correlations.

\subsection{Visual Odometry}
Similar to LiDAR odometry, visual odometry retrieves camera motion using information from images taken from different poses. Visual odometries can be simply divided into 2 classes: feature-based methods and direct methods.

\textbf{Feature-based methods.} These methods require feature extraction and association, mostly aiming at minimizing the reprojection error of the matched features. In feature extraction, typical image point features such as corners are well utilized, such as \cite{Howard2008}\cite{Kitt2010}. Line features and other novel features can also be used for different image scenes or camera sensors. \cite{Gomez-ojeda2016} combined ORB and LSD features to obtain more stable tracking in low-textured scenes. With multispectral cameras, \cite{Mouats2014} used log-Gabor wavelets to obtain interest points at different orientations and scales. In the optimization process, most works employ a nonlinear optimizer for feature matching between consecutive images, as previously mentioned \cite{Howard2008}\cite{Mouats2014}. In addition, some works exploit filtering methods to track the features over an image sequence. \cite{Kitt2010} used the iterated sigma point Kalman filter to track the ego-motion trajectory and feature observation. \cite{Zhang2012} considered image features as group targets and used the probability hypothesis density filter to track the group states. Most feature-based visual odometries share the same problem of computing efficiency and accuracy for data association, similar to feature-based LiDAR odometries. Moreover, feature-based visual odometries only concentrate on the features extracted without considering the information remaining in the images, which actually places a strong requirement on feature abundance.

\textbf{Direct methods.} To eliminate the shortcomings of feature-based visual odometries above, direct visual odometries have appeared in recent studies. They directly use camera sensor measurements without precomputation, considering the photometric error for pose estimation. For instance, \cite{Kerl2013} presented a direct method working for RGB-D cameras.
\cite{Engel2017} proposed direct sparse odometry, which combines photometric error minimization and the joint optimization of camera model parameters. \cite{Li2018} introduced a direct line guidance odometry, which uses lines to guide the key point selection.
There are also hybrid methods, such as semidirect visual odometry in \cite{Forster2014} and deep learning-based methods\cite{Wang2017deep}.

\subsection{Other Odometries}
To further improve the accuracy of the aforementioned odometries, many studies have attempted to incorporate inertial sensors. \cite{Tanskanen2015}\cite{Usenko2016}\cite{Qin2018} each propose a visual-inertial odometry. \cite{Hemann2016} used IMU to improve the performance of LiDAR odometry in long-range navigation. Second, using LiDAR and a camera together is another direction. \cite{Zhang2015} implemented visual-LiDAR odometry, which has better robustness in conditions of lacking visual features or aggressive motion. There are also some works using radar sensors\cite{Barjenbruch2015}.

\subsection{Error Model}
\label{subsec:Error Model}
As Table \ref{table:PBO} shows, only a small part of the ESO-related literature, such as \cite{Olson2009}\cite{Kitt2010}, presents error models for uncertainty estimation.
In these studies, the Hessian method and sampling method are 2 representative routes for error modeling.
Consider a simple odometry model as an example:
\begin{equation}
\label{eq:odomodel}
\hat\theta = \mathop{\arg\min}_{\theta}L(\theta|Z)
\end{equation}
where $\theta$ is the pose to be estimated as a column vector, $Z=\{z_i\}_{i=1,2...N}$ is the corresponding observation set consisting of $N$ sensor measurements with i.i.d. noise of variance $\sigma^2$, and $L$ is the objective function that measures the matching error between $\theta$ and $Z$. The error modeling methods can be formulated as follows.

\textbf{Hessian method.}
When $f$ is designed to be analytical and differentiable as
\begin{equation}
L(\theta)\triangleq \frac{1}{2}\sum_{i=1}^{N}\|z_i-h_i(\theta)\|_2^2
\end{equation}
where $h_i$ represents a measurement model mapping $\theta$ to $z_i$, Eq.\ref{eq:odomodel} can be solved using least squares.
Therefore, the close-formed solution of $\hat\theta_t$ can be approached recursively as
\begin{equation}
\label{eq:rls}
\begin{aligned}
\theta'&\leftarrow\theta+\delta\theta\\
\delta\theta &= -(\nabla^2_\theta L)^{-1}\nabla_\theta L
\end{aligned}
\end{equation}
in the Newton method,
so that the conditional covariance of $\hat{\theta}_t$ can be derived as
\begin{equation}
\label{eq:hessian}
Cov(\hat{\theta}_t|Z)=\sigma^2I{(\nabla^2_\theta L)}^{-1}
\end{equation}
where $\nabla^2_\theta f$ is defined as the Hessian matrix of $f$ in mathematics.
For the uncertainty of measurements, $Var(z_i)=\sigma^2$ is propagated to $\hat\theta$ by the inverse Hessian matrix of $L$, and this method is named the Hessian method.

Hessian methods are widely used in various odometries, such as \cite{Lu1997robot}\cite{Censi2007}\cite{Gomez-ojeda2016}\cite{Aksoy2014}.
However, several problems place restrictions on its usage.
First, for feature matching-based odometry, Hessian methods depend on a strong assumption that the feature correspondence $\{z_i, h_i\}$ is established correctly\cite{BrennaThesis2009}.
Second, in some cases, the inverse Hessian Matrix ${(\nabla^2_\theta L)}^{-1}$
is difficult to calculate but can be approximated from a Jacobian matrix such as \cite{Gomez-ojeda2016}\cite{Engel2013}, which actually decreases the covariance accuracy.
Third, it is difficult to ensure that step $\delta\theta$ in Eq.\ref{eq:rls} is infinitesimal, which is required by covariance calculation for nonlinear least squares.
Due to these problems, many studies, such as \cite{biber2003normal}\cite{Bengtsson2001}\cite{Bonnabel2016}, attempt to extend the Hessian method case by case, but it is still far from accurate.
 \begin{figure*}[ht]
 	\centering
 	\includegraphics[width=1.0\linewidth]{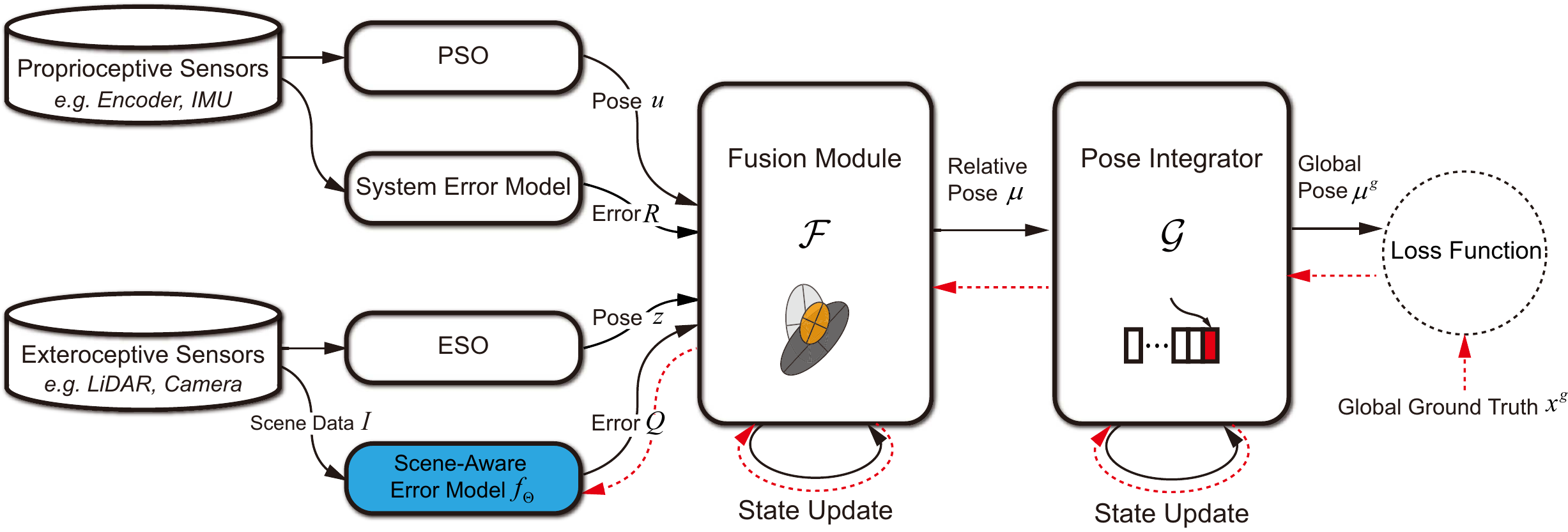}
 	\caption{Pipeline of scene-aware error modeling for fusion based localization.}
 	\label{fig:processingflow}
 \end{figure*}
 
\textbf{Sampling method.} For nonanalytical $L$, the covariance can be calculated by sampling poses according to a distribution $q$, such as the prediction from the motion model. Assuming the sample set $S_q=\{\theta^1,\theta^2,...,\theta^K\}$, the mean value of these samples can be regarded as an estimation of $\theta$
\begin{equation}
\label{eq:sampling}
\hat{\theta}=E_p(\theta)=\sum_{i=1}^{K}{p(Z|\theta_i)\theta_i}
\end{equation}
where $p$ denotes the probabilistic measurement model,
so that the covariance can be calculated as
\begin{equation}
Cov_p(\theta|S_q)=\sum_{i=1}^{K}{p(Z|\theta_i)(\theta_i-\hat{\theta})(\theta_i-\hat{\theta})^T}
\end{equation}
where the superscript $T$ represents the transpose operator here and later.

For instance, \cite{Kantor2002}\cite{Olson2009} used this method to calculate the covariance for scan matching. ROS\cite{quigley2009ros} package
\textit{amcl}\footnote{"amcl" is a probabilistic localization system for a robot moving in 2D, http://http://wiki.ros.org/amcl}
for 2D Monte Carlo localization also employs such a method to calculate covariance within a particle filter framework. In addition, there are some simulation methods\cite{Bengtsson2001}\cite{Buch2017} based on a prior environment map or geometry model.
In such methods, the sampling distribution $q$ and resolution have a great influence on the estimation results. The high computing cost also limits its online application.

Overall, existing error modeling methods strongly depend on a series of definitions and assumptions, which may have a negative influence on the uncertainty estimation. For example, the measurement model $h$ and objective function $L$ in the Hessian method and sampling distribution $p$ in the sampling method need to be manually designed or approximated, which may not objectively reflect the true relationship between target parameter $\theta$ and sensor observations $Z$. In addition, for fusion-based localization, there are another 2 important characteristics of ESO error that are often overlooked. First, the error model should be compatible with other localization modules for comparison. More importantly, the performance of ESO is sensitive to the scene.
Therefore, a scene-aware error model is needed to capture the relationship between odometry performance and the environment.

\section{Methodology}
\label{method}
\subsection{Fusion-based Localization with Scene-Aware Error Modeling}
Assume that a PSO such as dead reckoning has a system error, where the model parameters are calibrated and are not correlated with the scene.
Referring to such a PSO, a scene-aware error model of an ESO such as LiDAR or visual odometry can be learned, and a multimodal fusion-based localization can be achieved as described in Fig. \ref{fig:processingflow}.

At time $t$, let $u_t$ and $z_t$ be the relative poses estimated by PSO and ESO, respectively.
\begin{equation}
\label{eq:statetransition}
\left\{
\begin{aligned}
x_t &=u_t+\epsilon_t\qquad (\epsilon_t \sim R_t) \\
z_t&=x_t+\delta_t\qquad (\delta_t \sim Q_t) \\
\end{aligned}
\right.
\end{equation}
where $x_t$ is the true relative pose, $\epsilon_t$ and $\delta_t$ are Gaussian noise of $u_t$ and $z_t$ with their respective covariances $R_t$ and $Q_t$. $R_t$ is a system error, while $Q_t = f_{\Theta}(I_t)$ is predicted by a scene-aware error model $f_{\Theta}$ on data $I_t$ that describes the scene at the moment.

An information filter is used to find an MAP (maximum a posteriori) estimation of the relative pose $x_t$, which is represented by a mean pose $\mu_t$ and a covariance matrix $\Sigma_t$.
In Fig. \ref{fig:processingflow}, we denote the fusion module by $\mathcal{F}$, i.e., $\mu_t = \mathcal{F}(u_t,R_t,z_t,Q_t)$, which is operated recurrently and has the function of history memory. The process is detailed in the next section.

Since it is difficult to find accurate relative poses as the ground truth, we used the global pose by RTK-GPS instead. The supervision is conducted sporadically when the following two conditions are met: 1) a reliable GPS measurement is obtained, and 2) relative pose error has been accumulated for frames that exceed the error level of GPS.

Representing $\mu_t$ in a uniform matrix,
\begin{equation}
M(\mu_t) = \left[
\begin{matrix}
\mathbf{R} & \bf{t}  \\
\mathbf{0} & 1  \\
\end{matrix}
\right] %\tag{2}
\end{equation}
where $\mathbf{R,t}$ are the rotation matrix and translation vector, and the vehicle's mean pose $\mu^g_t$ at a global coordinate system can be estimated by accumulating the relative motions sequentially from an initial global pose $\mu_0^g$.
\begin{equation}
\label{uniformmatrix}
\begin{aligned}
M(\mu^g_t)&=M(\mu_0^g)M(\mu_1)M(\mu_2)...M(\mu_t)\\
&=M(\mu^g_{t-1})M(\mu_t)
\end{aligned}
\end{equation}
In Fig. \ref{fig:processingflow}, we denote the module of the pose accumulator by $\mathcal{G}$, i.e., $\mu^g_t = \mathcal{G}(\mu_t)$, which is operated recurrently and has the function of history memory.

When the supervision conditions are met, with a reliable GPS measurement $x^g_t$, the localization error $J(\mu^g_t,x^g_t)$ is backpropagated along the pipeline to optimize the parameters of the error model $f_{\Theta}$.
Hence, the major pipeline of fusion-based localization with scene-aware error modeling can be summarized by the following formulas, where for conciseness, subscript $t$ is omitted.

\begin{equation}
\label{eq:pipeline}
\begin{aligned}
Q &= f_{\Theta}(I)\\
\mu &= \mathcal{F}(u,R,z,Q)\\
\mu^g &= \mathcal{G}(\mu)\\
{\Theta}&=\mathop{\arg\min}_{\Theta}{J(\mu^g,x^g)}
\end{aligned}
\end{equation}

\subsection{PSO and ESO Fusion}

\subsubsection{Relative Pose Fusion Using an Information Filter}

The maximum a posterior probability (MAP) estimation of the vehicles’ relative motion state $x_t$ can be formulated as below, consisting of two subsequent steps in each iteration, i.e., prediction using vehicle control $u_t$ and updating using measurement $z_t$.
\begin{equation}
p(x_t|x_{t-1},u_t,z_t) \propto p(z_t|x_t) p(x_t|x_{t-1},u_t)
\end{equation}
When $x_t$ occurs in a very short time, this recursive estimation can also be regarded as a tracking process over a series of velocity measurements.

An information filter \cite{Thrun2005} can be used to estimate vehicle pose, with the multivariate Gaussian distribution represented using an information vector $\xi_t$ and an information matrix $\Omega_t$ in canonical representation as
\begin{equation}
\begin{aligned}
\xi_t&=\Sigma_t^{-1}\mu_t\\
\Omega_t &= \Sigma_t^{-1}
\end{aligned}
\end{equation}
where $\Sigma_t$ denotes the covariance matrix of this distribution. Obviously, $\Omega_t$ and $\Sigma_t$ are dual to measure the uncertainty.

By extending the original information filter, a relative pose fuser (RPF) is developed in this research, as listed in Algorithm \ref{if}.
At each frame, given $u_t$ and $z_t$ from the PSO and ESO modules, as well as their covariance matrices $R_t$ and $Q_t$ as uncertainty estimation,
rhe RPF function $\cal F$ estimates the mean relative motion $\mu_t$ at time $t$, and $\xi_t$ and $\Omega_t$.
Here, we use the information matrix $\tilde Q_{t}=Q_{t}^{-1}$ as the input of $\cal F$.
The reasons are two-fold: 1) taking $\tilde Q$ as an input of $\cal F$ can reduce the computing cost of the inversing matrix; and 2) the numerical estimation of $\tilde Q$ is more stable in the case of $\lVert Q\rVert\approx 0$ so that the formula of $\mu$ in Eq. \ref{eq:pipeline} should be rewritten as
\begin{equation}
\mu = \mathcal{F}(u,R,z,\tilde{Q})
\end{equation}
The original information filter and the derivation of Algorithm \ref{if} are given in Appendix \ref{appendix-i}.

\algdef{SE}[SUBALG]{Indent}{EndIndent}{}{\algorithmicend\ }%

\algtext*{Indent}

\algtext*{EndIndent}

\begin{algorithm}[h]
	
	\caption{RPF ${\cal{F}}(u_t,R_t,z_t,\tilde Q_{t})$}
	
	\label{if}
	
	\begin{algorithmic}[1]
		\label{algo:LPF}
		\State \bf{Coordinate transformation}\normalfont$(\xi_{t-1},\Omega_{t-1})$
		\Indent
		$\left\{
		\begin{aligned}
		\hat{\xi}_{t-1}&=\mathbf{0}\\	 	
		\hat{\Omega}_{t-1}& = M(\mu^{-\theta}_{t-1})\Omega_{t-1}M^T(\mu^{-\theta}_{t-1})
		\end{aligned}
		\right.$
		\EndIndent
		\State ${\overline{\Omega}_t = (\hat{\Omega}_{t-1}^{-1}+R_t)^{-1}}$
		\State $\overline{\xi}_t=\overline{\Omega}_t(\hat{\Omega}_{t-1}^{-1}\xi_{t-1}+u_t)$%=\overline{\Omega}_tu_t$
		\State $\Omega_t=\tilde Q_{t}+\overline{\Omega}_t$
		\State $\xi_t=\tilde Q_{t}z_t+\overline{\xi}_t$
		\State $\mu_t =\Omega_t^{-1}\xi_t $	
		\State \bf{Output:\em}\normalfont {$\mu_t$}%\normalfont {$\xi_{t},\Omega_{t},\mu_t$}
	\end{algorithmic}
\end{algorithm}

Step 1 of Algorithm \ref{if} is coordinate transformation. Since this research estimates the vehicle's relative pose using the information filter, at each frame $t$, we have $\hat{\mu}_{t-1}=\mathbf{0}$ denoting the zero position and consequently $\hat{\xi}_{t-1}=\Sigma_{t-1}^{-1}\hat{\mu}_{t-1}=\mathbf{0}$, where $\mathbf{0}$ is used to represent the zero vector or matrix here and later.
On the other hand, the information matrix $\Omega_{t-1}$ obtained in the previous iteration needs to be transformed to compensate for the rotation factor in $\mu_{t-1}$. Assume $\mu_{t-1}$ can be decoupled as $(\mu_{\mathbf{\theta},t-1},\mu_{\mathbf{t},t-1})$, where $\mu_{\mathbf{\theta},t-1}$ denotes the rotation factor and $\mu_{\mathbf{t},t-1}$ is the translation factor.
Let $\mu^{-\theta}_{t-1}\leftarrow(-\mu_{\mathbf{\theta},t-1}, \mathbf{0})$ we have
\begin{equation}
\hat{\Omega}_{t-1} = M(\mu^{-\theta}_{t-1})\Omega_{t-1}M^T(\mu^{-\theta}_{t-1})
\end{equation}

Here, $\mu_{t-1}$, $\xi_{t-1}$ and $\Omega_{t-1}$ denote the results obtained in the estimation of time $t-1$, which are relative to the vehicle's coordinate system at time $t-2$. Whereas $\hat\mu_{t-1}$, $\hat\xi_{t-1}$ and $\hat\Omega_{t-1}$ are the converted results in the estimation of time $t$, which are relative to the vehicle's coordinate system at time $t-1$.

Steps 2-3 predict the information matrix $\overline{\Omega}_t$ and vector $\overline{\xi}_t$ by incorporating the outputs of the referred PSO module, steps 4-5 are measurement updating using results from the target ESO, and step 6 is conversion from canonical representation to find a mean relative pose $\mu_{t}$.
\begin{figure*}[!h]
	\centering
	\includegraphics[width=1.0\linewidth]{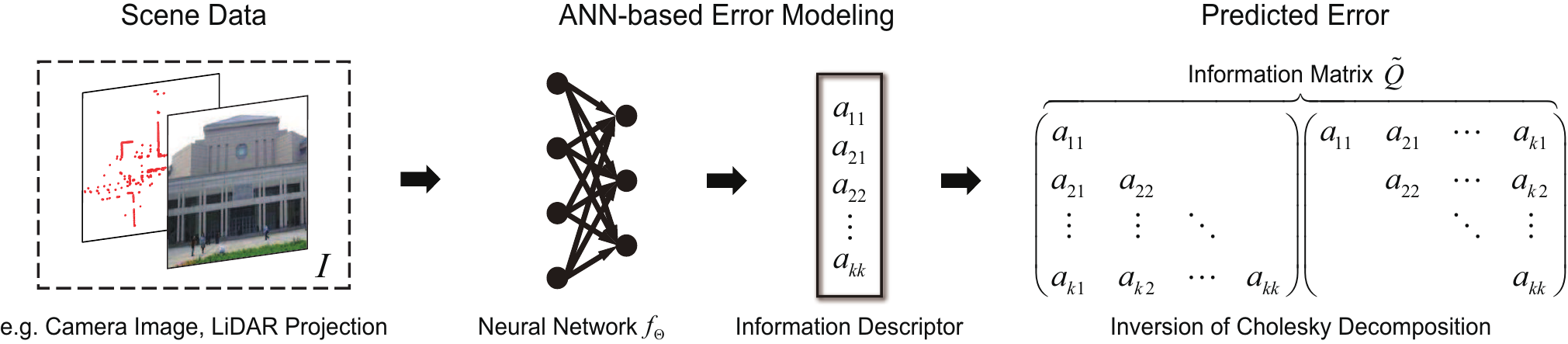}
	\caption{Scene-aware error modeling, mapping from scene data to predicted error using information matrix.}
	\label{fig:covmappingsimple}
\end{figure*}
\subsubsection{Extension to Multimodal Fusion}
This model can be easily extended to a system with other mutually independent odometry modules. The output $z_i$ from the $i$th target ESO module can be seen as independent observation variables similar to the 2-module system. Assuming that there are $N$ target ESO modules, the $i$th observation and its covariance are $z_i,Q_i$, the probabilistic formulation can be extended as
\begin{equation}
\begin{aligned}
p(x_t|x_{t-1}&,u_t,z_{1,t},z_{2,t}...,z_{n,t}) \propto \\
p&\left(x_t|x_{t-1},u_t\right)\prod_{i=1}^n p\left(z_{i,t}|x_t\right)
\end{aligned}
\end{equation}
so that in the information filter framework, steps 4 and 5 in Algorithm \ref{if} can be rewritten as

\begin{equation}
\begin{aligned}
&\Omega_t=\overline{\Omega}_t+\sum_{i=1}^n\tilde Q_{i,t}\\
&\xi_t=\overline{\xi}_t+\sum_{i=1}^n\tilde Q_{i,t}z_{i,t}
\end{aligned}
\end{equation}

\subsection{Scene-Aware Error Model Learning}
\label{subsec:learning}

\subsubsection{Error Modeling}

The pipeline is shown in Fig. \ref{fig:covmappingsimple}. For ESO, such as LiDAR/visual odometry, the scene data can be obtained from its input, such as a camera image or LiDAR point cloud, and the next step is to map it to the pose error, namely, the information matrix in RPF.
An information matrix $\tilde Q$ is symmetric and positive definite when $\tilde Q\not=\mathbf{0}$; hence, it can be factorized by Cholesky decomposition
\begin{equation}
\tilde Q = LL^T
\end{equation}
where
\begin{equation}
L = \left[
\begin{matrix}
a_{11} &  &  &\\
a_{21} & a_{22} & & \\
\vdots& \vdots& \ddots& \\
a_{k1} & a_{k2} & \cdots &a_{kk}
\end{matrix}
\right]
\end{equation}
is a unique lower-triangular matrix of $\tilde Q$. Define an information descriptor $\mathbf{a}=(a_{11}, a_{21},...,a_{kk})^T$ consisting of all the independent elements of $L$. The neural network needs to be customized for different scene information $I$ and input output to $\mathbf{a}$, with which $\tilde Q$ can be uniquely estimated as the predicted error.

\subsubsection{Loss Function}

To learn a parameter set $\Theta$ of the neural network $f_\Theta$ in Fig. \ref{fig:covmappingsimple}, supervised learning is not adaptive, as the ground truth of neither the information descriptor $\mathbf{a}$ nor the information matrix $\tilde Q$ is available.
However, the vehicle's ground truth position can be obtained under certain conditions.
For example, a vehicle pose $x^g_t$ can be measured using, e.g., a GPS/IMU suite or a loop closure detector.
$x^g_t$ is considered as a ground truth location if and only if
\begin{equation}
\lVert x^g_t-\mu^g_t\rVert >> e
\label{gtcondition}
\end{equation}
where $\mu^g_t$ is the estimation at the time by fusing scan matching and dead reckoning outputs, and
$e$ is the error level of the measurement $x^g_t$.

Given a parameter set $\Theta$, the localization module initiates from a ground truth $x^g_0$ and estimates vehicle pose $\left\{\mu_{0}^{g},\mu_{1}^{g}...,\mu_{T}^{g} \right\}$ for $T$ steps.
The localization error caused by $\Theta$ is accumulated during these steps, which is evaluated at time $T$ as below, where $\lambda$ is a hyperparameter to weight errors in location and heading angle.
\begin{equation}
\label{lossfunc}
J(\mu_{T}^{g},x_{T}^{g})=\left\| \mu_{T}^{g} - x_{T}^{g}\right\|^2_2+\lambda \left|\mu_{\theta,T}^{g} - x_{\theta,T}^{g}\right|^2
\end{equation}

Given a pair of ground truth positions $\{x^g_0,x^g_T\}$, the objective is to learn a parameter set $\Theta$ to minimize error $J$, subject to $\mu_{0}^{g}=x^g_0$.

\begin{equation}
\label{lossfunc_condition}
\begin{aligned}
{\bf min.}  & J(\mu_{T}^{g},x_{T}^{g})\\
{\bf s.t.}  & \mu_{0}^{g}=x^g_0\\
\end{aligned}
\end{equation}

\subsubsection{Parameter Learning}
\begin{algorithm}[!h]
	\caption{Learning Error Model of ESO}
	\label{SMCL}
	\begin{algorithmic}[1]
		\Require 
		Initialize the parameters $\Theta$ in $f$. Set maximum iteration
		$M$, $i=0$
		\While{$i<M$}
		\State {\it Reset:}
		\State {Initialize starting point $\mu_0^g=x_0^g$.}
		\State {Initialize RPF $\cal F$ .}
		\State {\it Forward Prediction:}
		\For{$t=1$ to $T$}
		\State Get ${(u_t, R_t)}$ from $\mathcal{L}_{ref}, f_{ref}$
		\State Get $z_t$ from $\mathcal{L}_{tar}$
		\State $f_{\Theta}(I_t) \rightarrow \tilde Q_t$
		\State $\mu_t={\cal F}(u_t,R_t,z_t,\tilde Q_t)$
		\State $\mu_{t}^{g}={\cal G}(\mu_{t-1}^{g}, \mu_t)$
		\EndFor
		\State {\it Back Propagation:}
		\For{$t=T$ to $1$}
		\If{$1\leq t<T$}
		\State $\nabla J_{\tilde{Q}_t} =
		\frac{\partial J}{\partial u_t}\frac{\partial u_t}{\partial\tilde{Q}_t}+
		\frac{\partial J}{\partial \Omega_t}\frac{\partial \Omega_t}{\partial\tilde{Q}_t}+
		\frac{\partial J}{\partial \xi_t}\frac{\partial \xi_t}{\partial\tilde{Q}_t}$
		\Else
		\State {$\nabla J_{\tilde{Q}_t} = \frac{\partial J}{\partial u_t}\frac{\partial u_t}{\partial\tilde{Q}_t}$}
		\EndIf
		\EndFor
		\State $\nabla J_\Theta= \sum_{t=1}^{T}\nabla J_{\tilde{Q}_t}\frac{\partial\tilde{Q}_t}{\partial\Theta}$
		\State $\Theta \leftarrow　\Theta　- \nabla J_\Theta$
		\State $i=i+1$
		\EndWhile
	\end{algorithmic}
\end{algorithm}
\begin{figure*}[!htbp]
	\centering
	\includegraphics[width=1.0\linewidth]{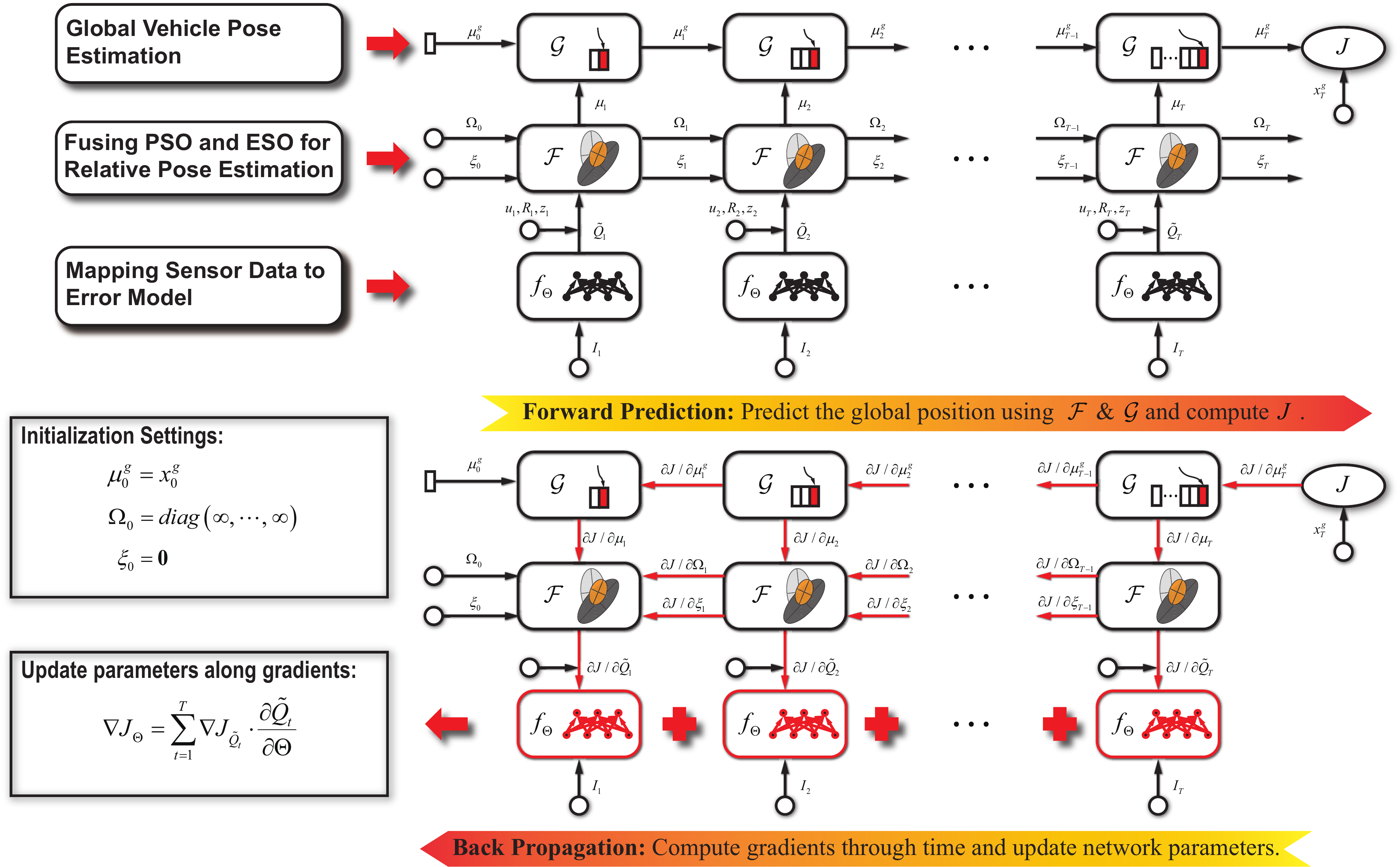}
	\caption{Training pipeline. Forward prediction: estimate a sequence of vehicle pose on the current parameter set $\Theta$ for $T$ steps. Back propagation: optimize $\Theta$ to minimize the error $J$ between the estimated vehicle pose $\mu_{T}^{g}$ at time $T$ with its ground truth $x_{T}^{g}$.}
	\label{fig:3 net expansion}
\end{figure*}
$\Theta$ is refined iteratively whenever a pair of the vehicle's ground truth position $<x^g_0,x^g_T>$ is obtained, where learning is conducted in two subsequent steps: forward prediction and backpropagation, which are described in Fig. \ref{fig:3 net expansion} and Algorithm \ref{SMCL}.

Forward prediction estimates a sequence of vehicle poses on the current parameter set $\Theta$ for $T$ steps, where the process of each step is described in lines 5-13 of Algorithm \ref{SMCL}. Initiated from $\mu_{0}^{g}=x^g_0$, forward prediction results in an estimation of the vehicle pose $\mu_{T}^{g}$ at time $T$.

Backpropagation refines $\Theta$ to minimize the error $J$ between the estimated vehicle pose $\mu_{T}^{g}$ at time $T$ and its ground truth $x_{T}^{g})$. The functions $\cal F$, $\cal G$, and the neural network $f_\Theta$ are differentiable, and the error $J$ can be backpropagated from time $T$ to $1$ by stochastic gradient descent. The gradient estimation and the backpropagation process are described in lines 14-23 of Algorithm \ref{SMCL}.

\section{Experiment on LiDAR Odometry}
\label{experiment}
\begin{figure}[!h]
	\centering
	\includegraphics[width=1.0\linewidth]{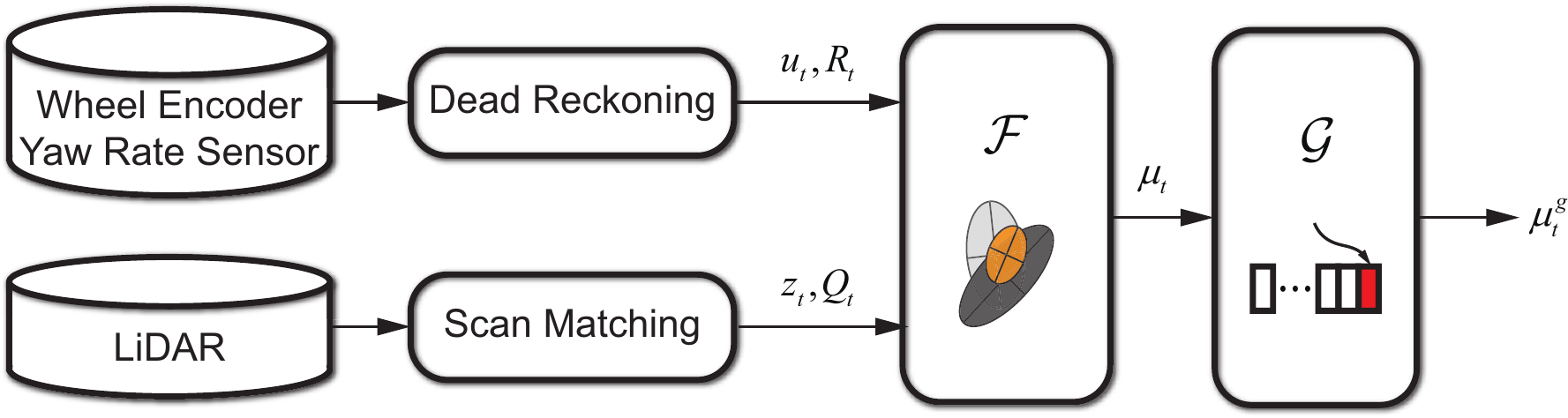}
	\caption{Processing flow of experiment on LiDAR odometry.}
	\label{fig:processingflow-lo}
\end{figure}
An overview of the processing flow for fusing LiDAR odometry is given in Fig. \ref{fig:processingflow-lo}. A simple dead reckoning (DR) is used as the referred PSO odometry. For the target LiDAR odometry, two classical 2D scan matching algorithms, CSM\cite{Olson2009} and PLICP\cite{Censi2008}, are selected to perform error model learning, and their traditional error models \cite{Olson2009} and \cite{Censi2007} are used to compare with our method, corresponding to the aforementioned sampling method and Hessian method, respectively.

Three experimental results are presented. First, simulation data at specifically designed simulation environments are used to verify the proposed method and demonstrate that the predicted error models can capture scene properties. Second, real-world data from an instrumented vehicle are used, where training and testing are conducted in the same campus environment to compare the performance. Third, experiments in an unexperienced environment are conducted, where training and testing are performed at different sites to demonstrate the generality of the proposed method in unexperienced scenes.

\subsection{State Definition and Network Design}
For intelligent vehicles, generally, 2-dimensional localization is sufficient in the structural urban environment, so that we set $x_t,u_t,z_t\in SE(2)$ in Eq. \ref{eq:statetransition}. More specifically, for relative positioning, any pose state is defined as a column vector including 3 independent elements $(\Delta x, \Delta y, \Delta \theta)$, where $\Delta x, \Delta y$ are the displacement relative to the zero position in the local coordinate system, and $\Delta \theta$ is the heading change in the Euler angle.

\begin{figure}
	\centering
	\includegraphics[width=0.9\linewidth]{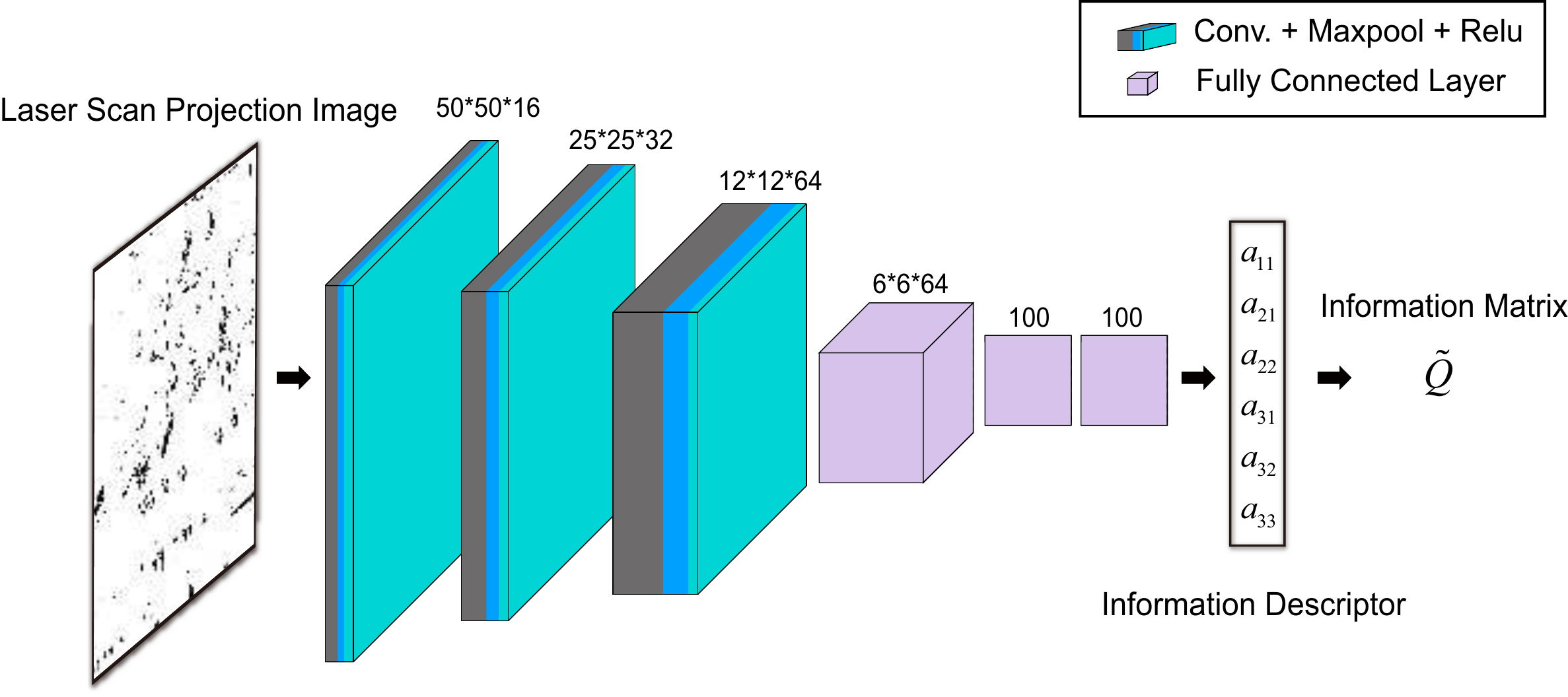}
	\caption{Scene-aware error modeling for LiDAR odometry.}
	\label{fig:network-eng}
\end{figure}

For local relative localization, there is no considerable scenario change when cars move such a short distance (several meters). Therefore, only one of the two frames in scan matching is enough to represent the local scenario, which contains sufficient scene information as network input.
Therefore, given a LiDAR scan $P$, a neural network $f_\Theta$ is designed to map it to an information matrix $\tilde Q$ that models the error of scan matching result $z$ on $P$.
A CNN (convolutional neural network)\cite{krizhevsky2012imagenet} is used due to its superior performance, which has been demonstrated in the literature such as \cite{ren2015faster}\cite{sharif2014cnn}\cite{gidaris2015object}.
Therefore, a LiDAR scan $P$ is first converted to a binary image by regularly tessellating an ego-centered horizontal space, and each pixel value is $0$ or $1$ in which $0$ means there is no LiDAR observation falling into the grid and $1$ indicates that at least one LiDAR beam hit the grid. In this research, considering learning efficiency and the sparsity of LiDAR points, a $50\times50$ image is generated for each scan at a dimension of $120m\times120m$, and the pixel size is $2.4m\times2.4m$. The detailed network structure is given in Fig. \ref{fig:network-eng}.

\subsection{Simulation Data Experiment}

\begin{figure}[!h]
	\centering
	\includegraphics[width=0.99\linewidth]{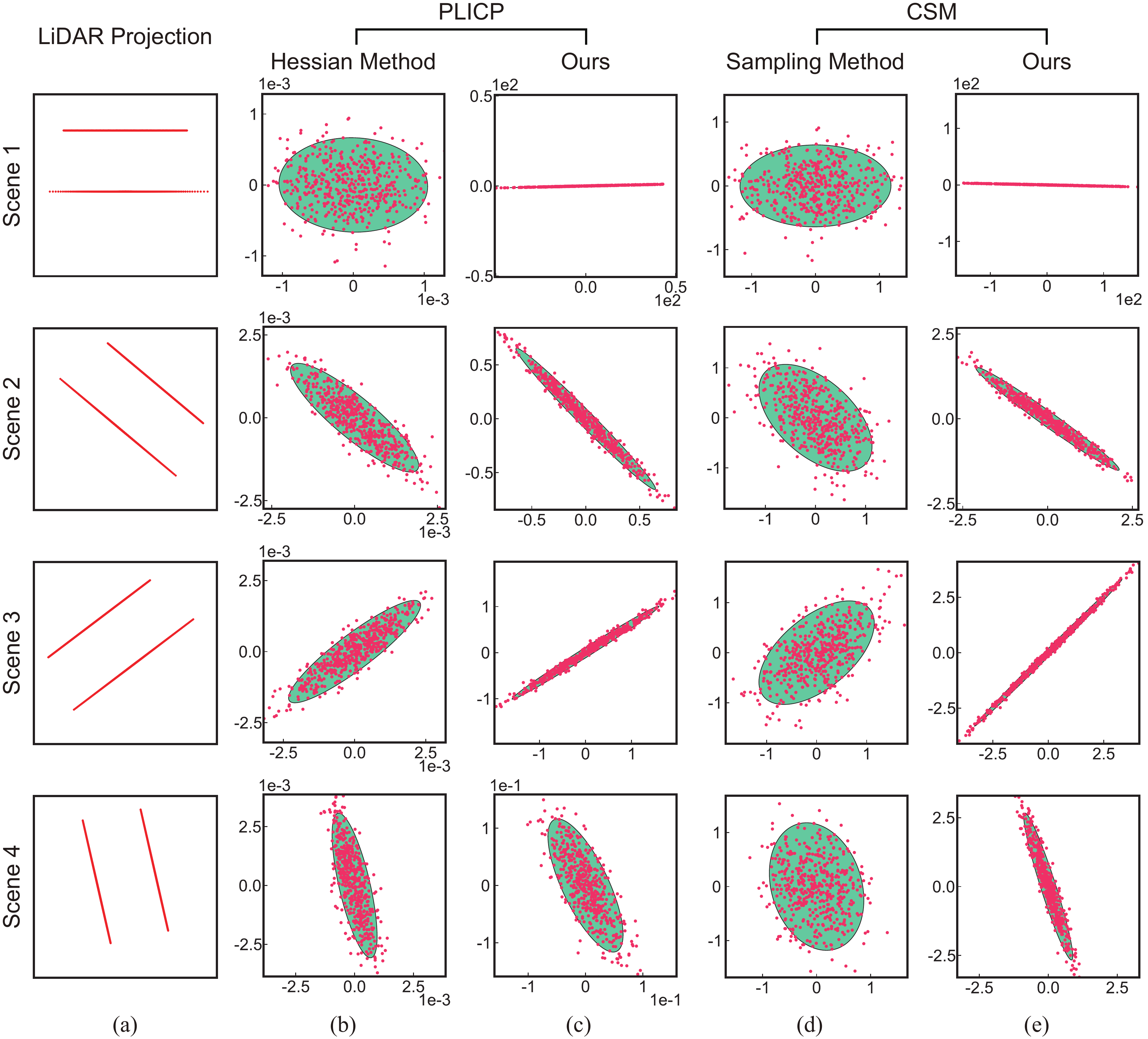}
	\caption{The simulated scenes and learning results of scene-aware error model. (a)LiDAR scans; (b)Error modeling of PLICP using Hessian method; (d)Error modeling of CSM using sampling method; (c)(e)our learned error model. The oval in (b)-(e) denotes the 2 standard deviation and the points are the sampling results from the Gaussian covariance. 
	}
	\label{fig:covinstances_corridor}
\end{figure}

\subsubsection{Dataset}
Gazebo\cite{koenig2004design} and ROS\cite{quigley2009ros} are used as the simulator to build an artificial environment and collect simulated sensor data.
As corridor scenes with two parallel featureless walls are very challenging for scan matching and their fusion-based approaches, such an environment is built, as shown in the first column of Fig. \ref{fig:covinstances_corridor}.
The sensor set of the simulated car model includes a 360-degree horizontal LiDAR for scan matching, a wheel encoder, and a yaw rate sensor for dead reckoning. To make the simulated data more realistic, Gaussian noise is added to these sensor readings.

In data collection, the simulated car traveled along the corridor with a series of steering operations so that the direction of the LiDAR frame changed continuously. Two sets of data are collected for training and testing by driving the car along a rectangular and a circular trajectory, respectively.

\subsubsection{Learning Result of Scene-Aware Error Model}
Because the corridor walls are straight and parallel, the point features are monotonous. The error distribution of scan matching at such a scene usually has a main direction along the direction of the passage. Moreover, the covariance can be estimated by a conventional solution \cite{Olson2009}\cite{Censi2007} for comparison. Several typical cases in the testing process are shown in Fig. \ref{fig:covinstances_corridor}, where the results of our method are compared with those of a conventional method that are shown side by side. The covariances are represented by 2 standard deviation ovals and sampled scattered points, which are drawn in the ${x-y}$ plane.
Apparently, due to the dependence of sampling, although the conventional solution can give the correct main direction of covariance, the scale is not accurate enough, which may lead to a worse localization fusion.
\begin{figure} 
	\centering
	\subfloat[Euclidean distance error.\label{fig:simulation-acc-a}]{%
		\includegraphics[width=0.49\linewidth]{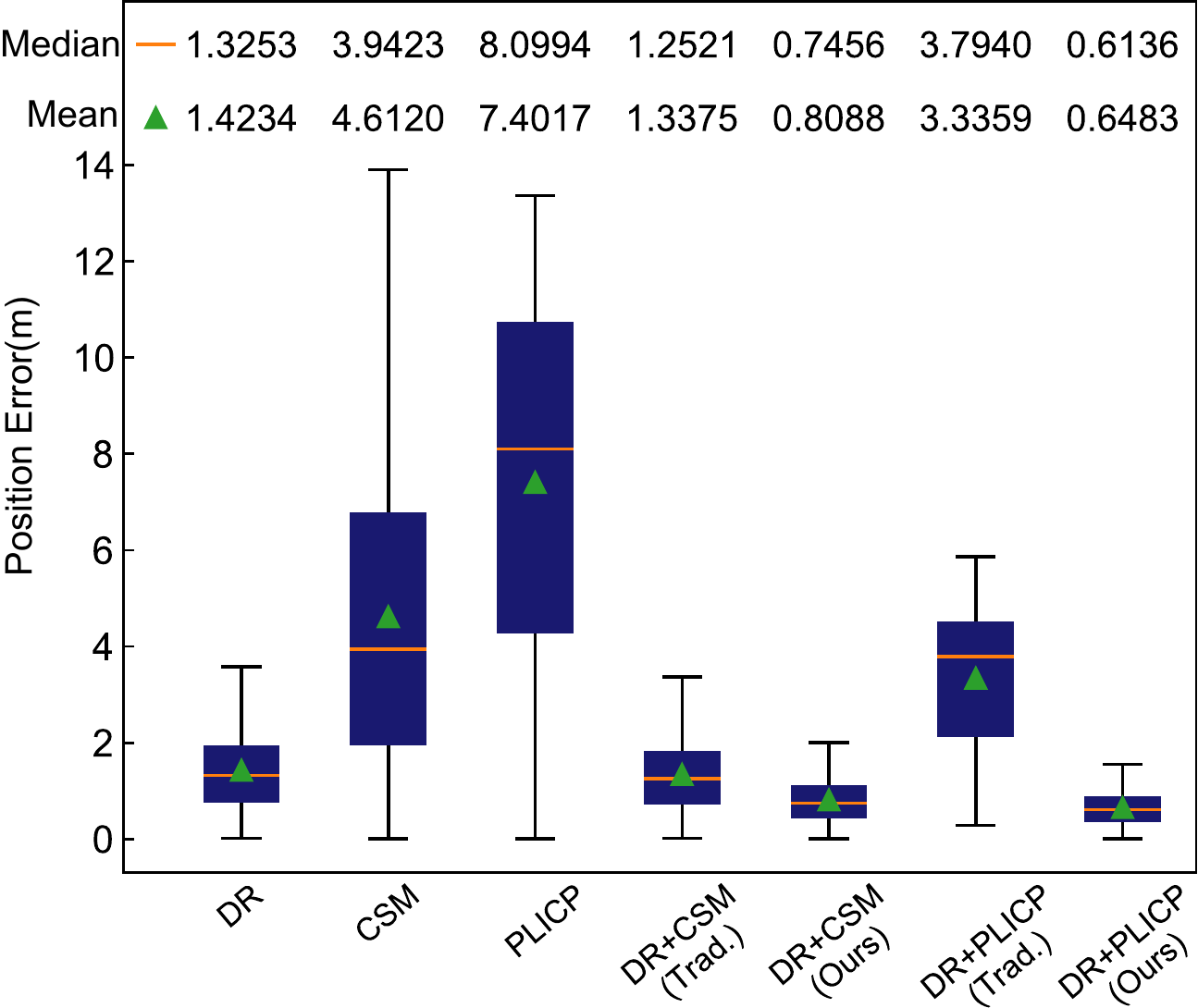}}
	\hfill
	\subfloat[Heading error.\label{fig:simulation-acc-b}]{%
		\includegraphics[width=0.48\linewidth]{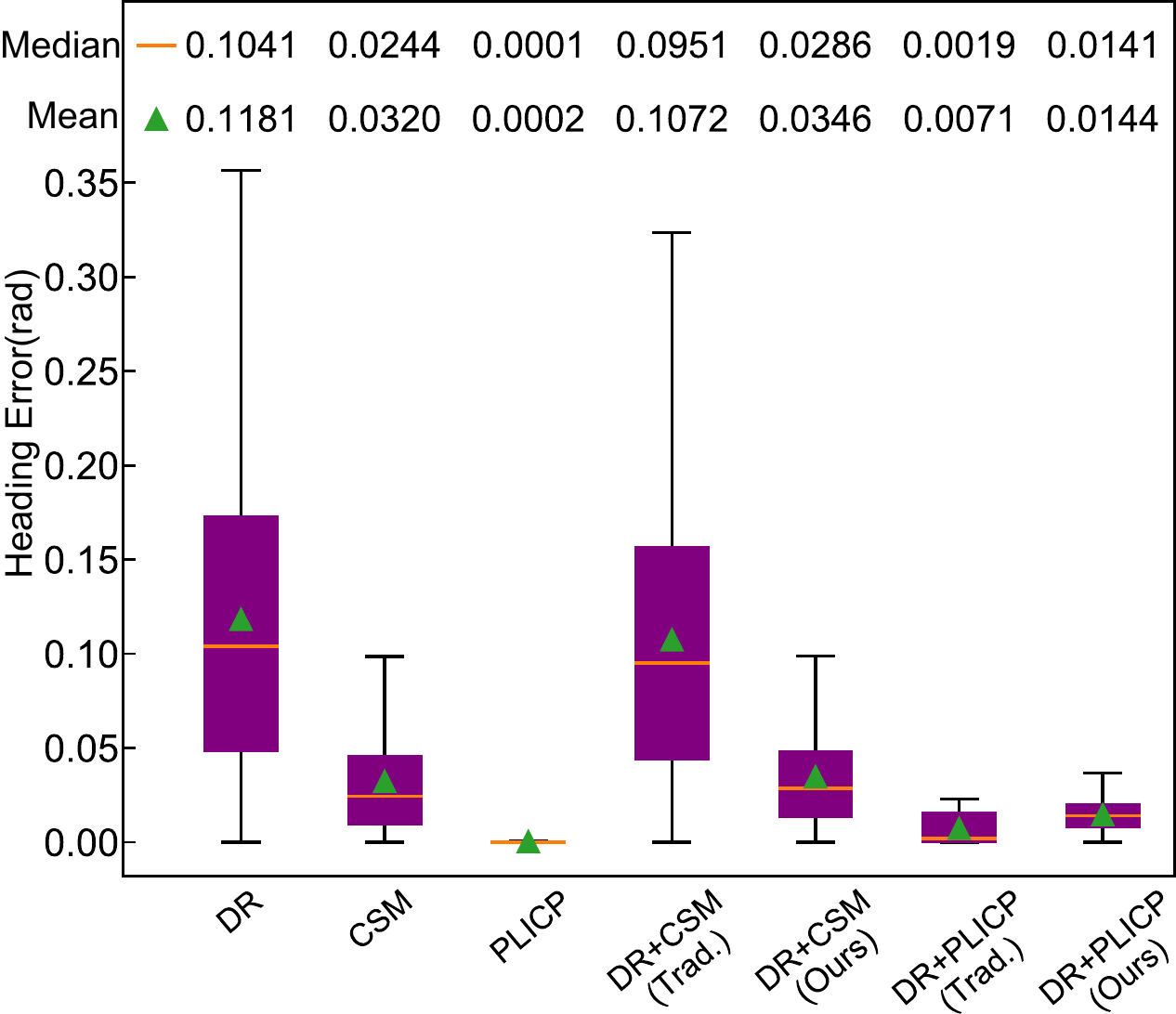}}
	\caption{Localization accuracy using LiDAR odometry on simulation data.}
	\label{fig:simulation-acc} 
\end{figure}

\begin{figure*}[!h]
	\centering
		\label{fig:train-epoch-change}
	\subfloat[Learning results of scene-aware error model.]{%
		\begin{minipage}[c][2in]{4.5in}
			\centering
			\label{fig:train-epoch-change-a}
			\includegraphics[width=4.5in]{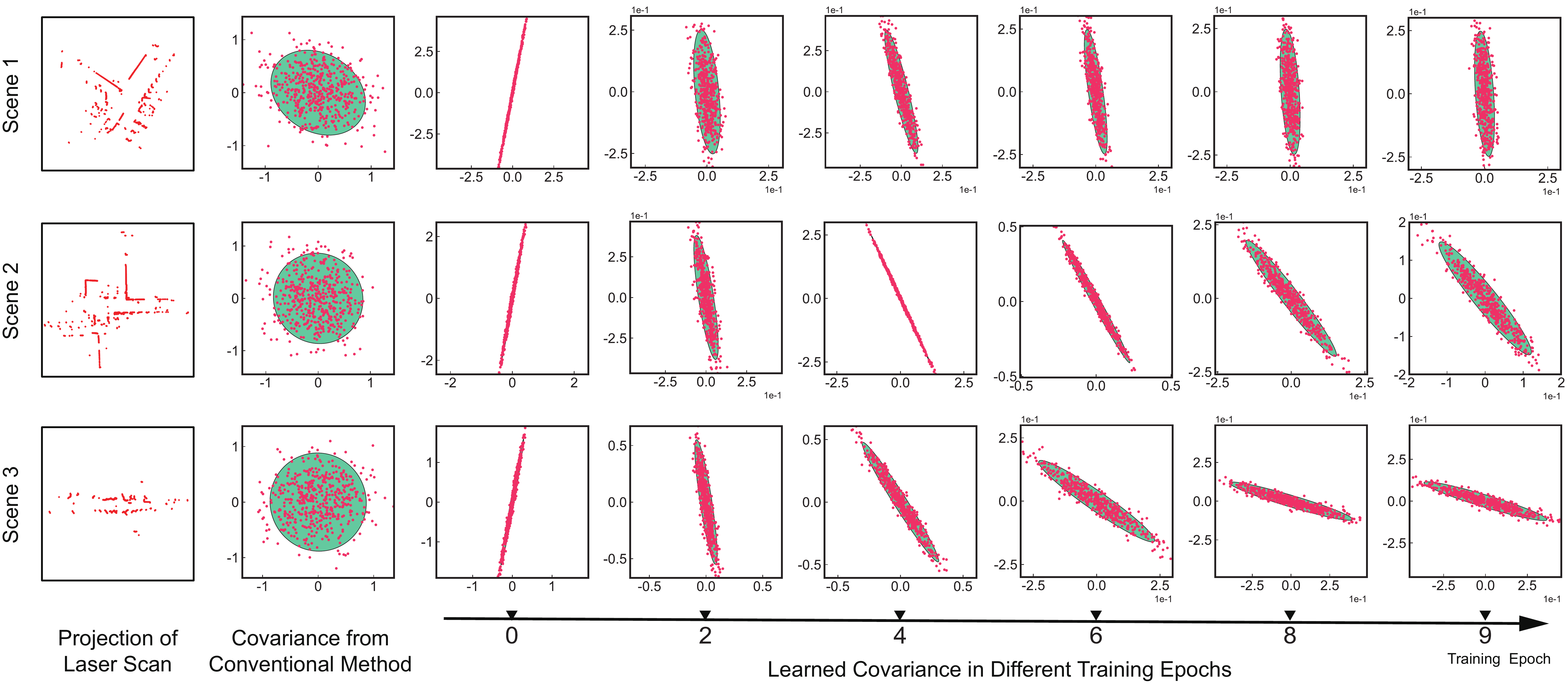}
		\end{minipage}}%
		\hspace{\tabcolsep}
		\subfloat[Trajectories comparison.]{%
			\begin{minipage}[c][2in]{2in}
							\label{fig:train-epoch-change-b}
				\centering
				\includegraphics[width=2in]{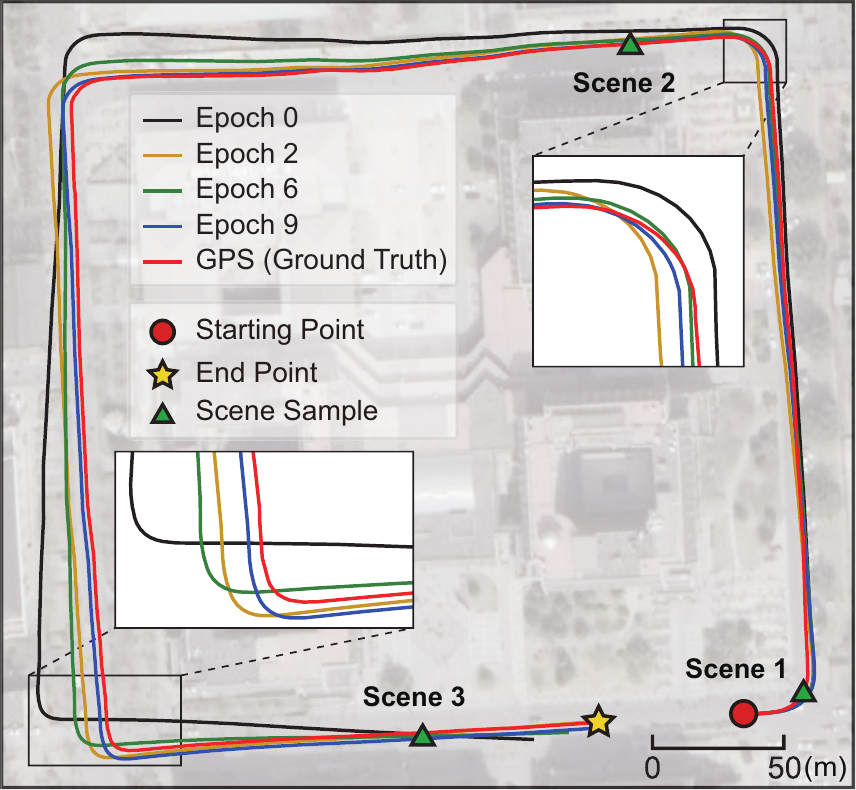}
			\end{minipage}}
			\caption{Learning results of LiDAR odometry(CSM\cite{Olson2009}) on real data. (a)Real scenes and error modeling results of CSM at each training epoch; (b)The global localization results(DR+CSM) using the error model of each training epoch.}
\end{figure*}
\subsubsection{Localization Accuracy}
In contrast, our model can calculate more than the correct main direction of covariance, and it also obtains a more accurate covariance scale for fusion, which matches the error scale of odometry. The position error statistics of every 40-meter-long trajectory segment in the testing process are shown in Fig. \ref{fig:simulation-acc}, in which Fig. \ref{fig:simulation-acc-a} gives the Euclidean distance error distribution at the end of every trajectory, and Fig. \ref{fig:simulation-acc-b} shows the corresponding yaw error distribution. Our method has obvious advantages in the comparison of localization accuracy on both of these LiDAR odometry algorithms.
\begin{figure}[!h]
	\centering
	\includegraphics[width=0.7\linewidth]{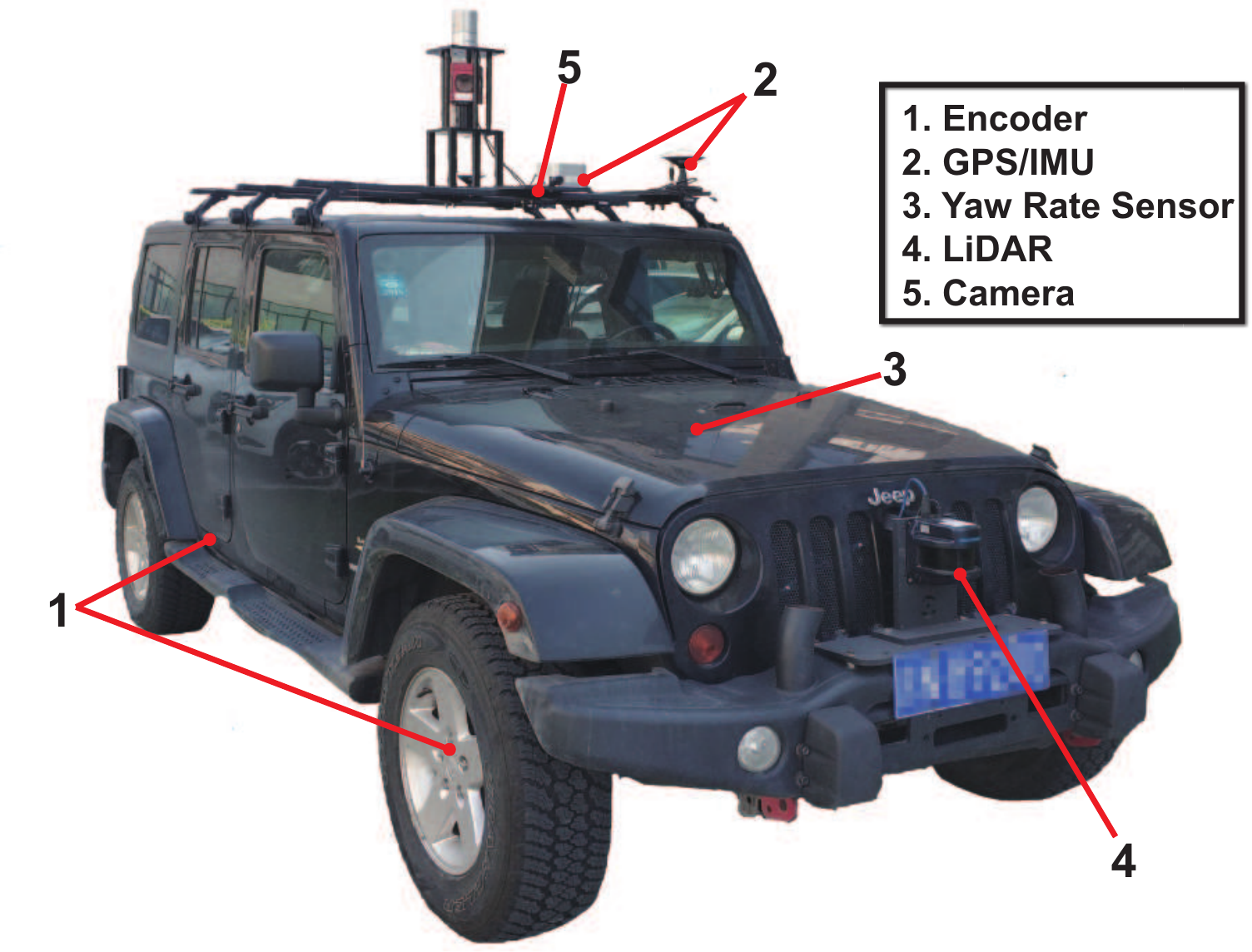}
	\caption{The instrumented vehicle used in real data collection.}
	\label{fig:jeepcar}
\end{figure}
\subsection{Real Data Experiment}

\subsubsection{Dataset}
An instrumented vehicle, as shown in Fig. \ref{fig:jeepcar}, is used to collect data at a real-world scene to evaluate the performance of the proposed method. The following sensors are used: 1) LiDARs are horizontally mounted in the front and rear of the car profile for scan matching; 2) a wheel encoder and a yaw rate sensor are used for dead reckoning; and 3) a highly accurate GPS/IMU suite is used to obtain ground truth locations of the vehicle for model training and localization result evaluation.

For experiments on experienced scenes, two sets of data are collected in the same region of the Peking University campus for training and testing, which are conducted on different days. For experiments on unexperienced scenes, we collect a large-scale dataset in several different regions with a total mileage of approximately 10 km. Testing data accounts for 40\% of the dataset, most of which cannot be seen by LiDAR in the training data. To avoid the great accuracy disparity between the referred PSO and the target ESO, which may lead to no complementary information for fusion, we manually adjusted the accuracy settings of the sensors in the experiments for different control groups.
\begin{figure}[!h]
	\centering
	\subfloat[Trajectory comparison of different methods.]{%
		\label{fig:traj-experienced}
		\includegraphics[width=1.0\linewidth]{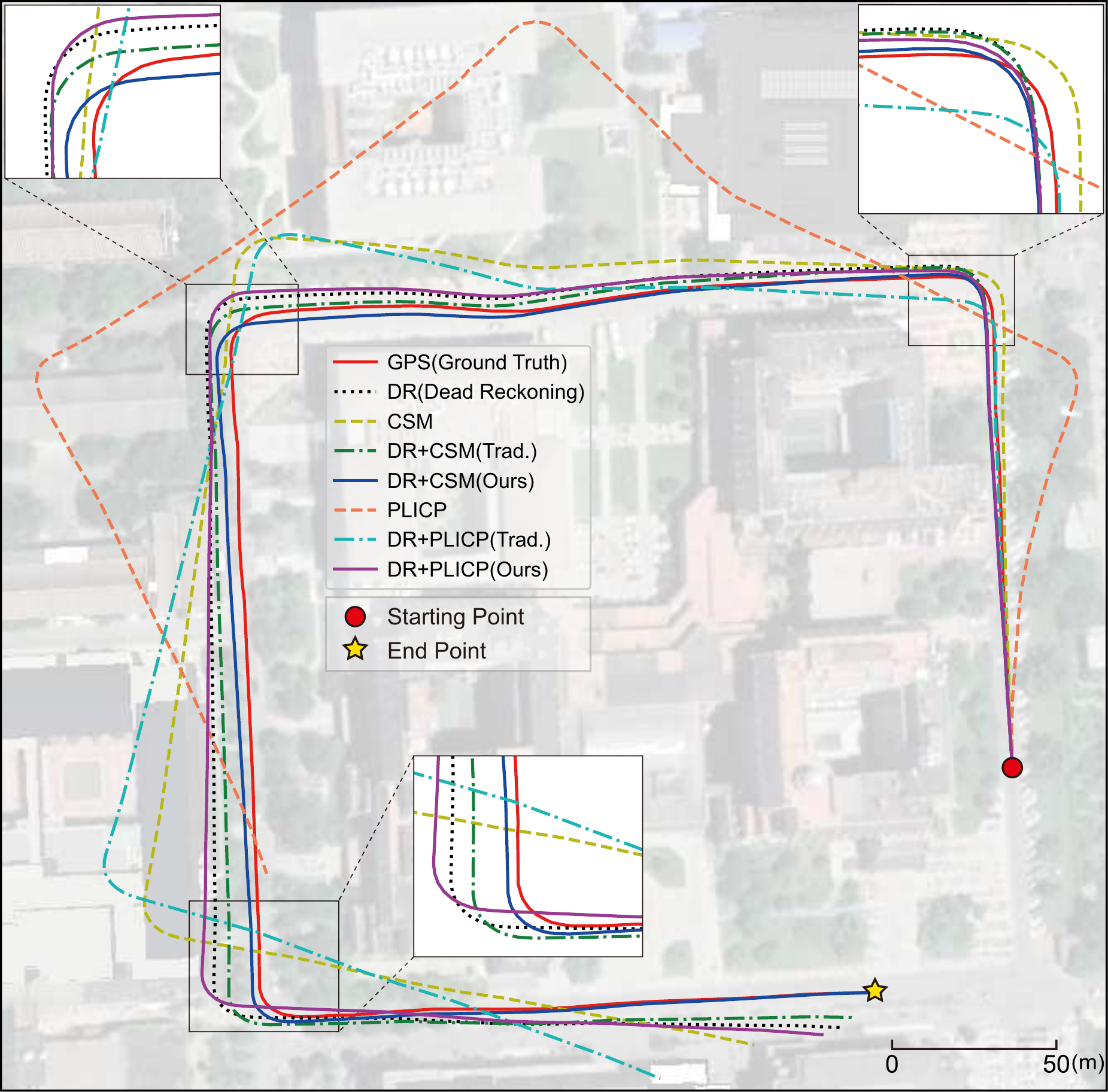}}
	
	\subfloat[Euclidean distance error.]{%
		\label{fig:accdis-experienced}
		\includegraphics[width=0.49\linewidth]{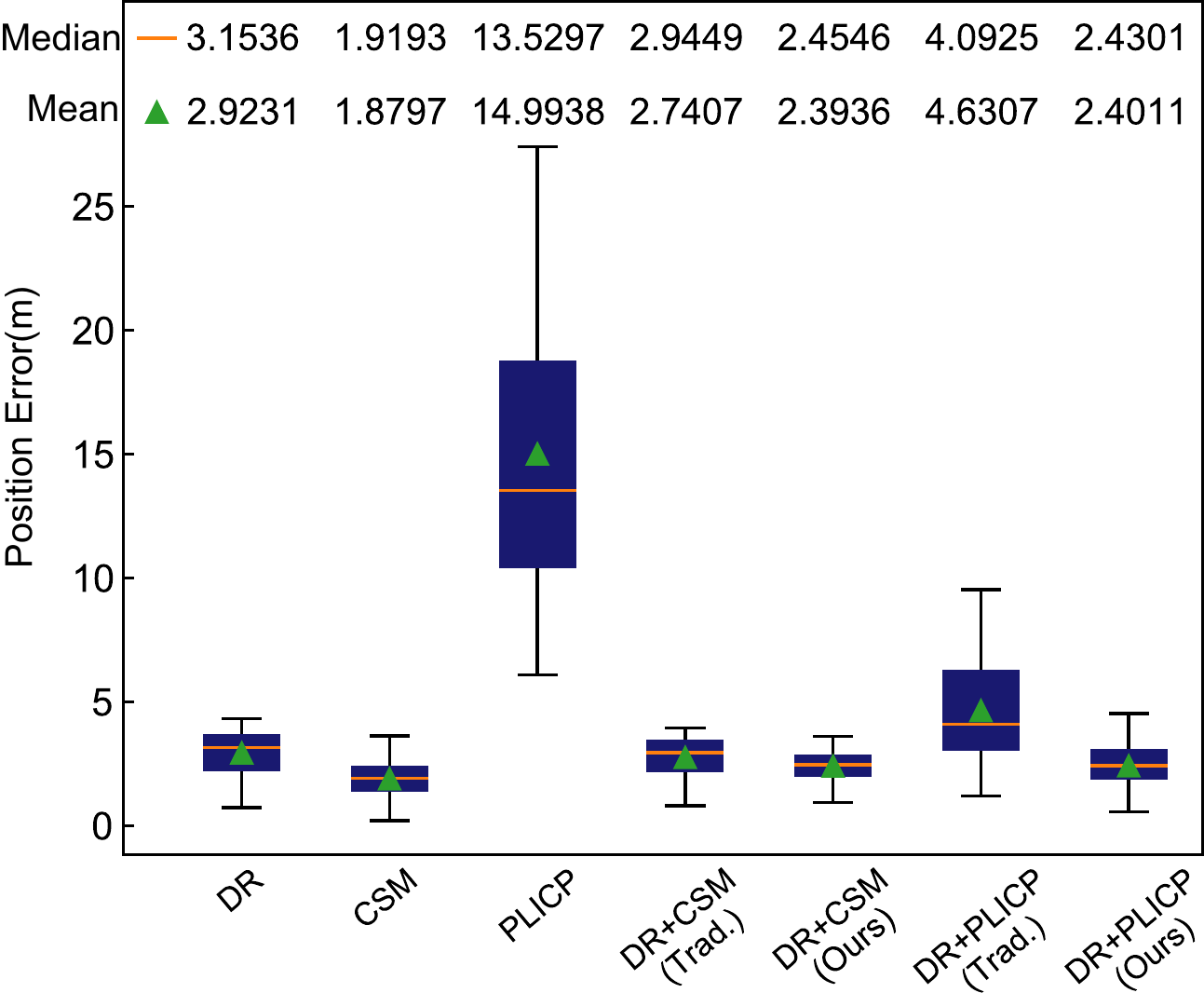}}\hfill
	\subfloat[Heading error.]{%
		\label{fig:accyaw-experienced}
		\includegraphics[width=0.48\linewidth]{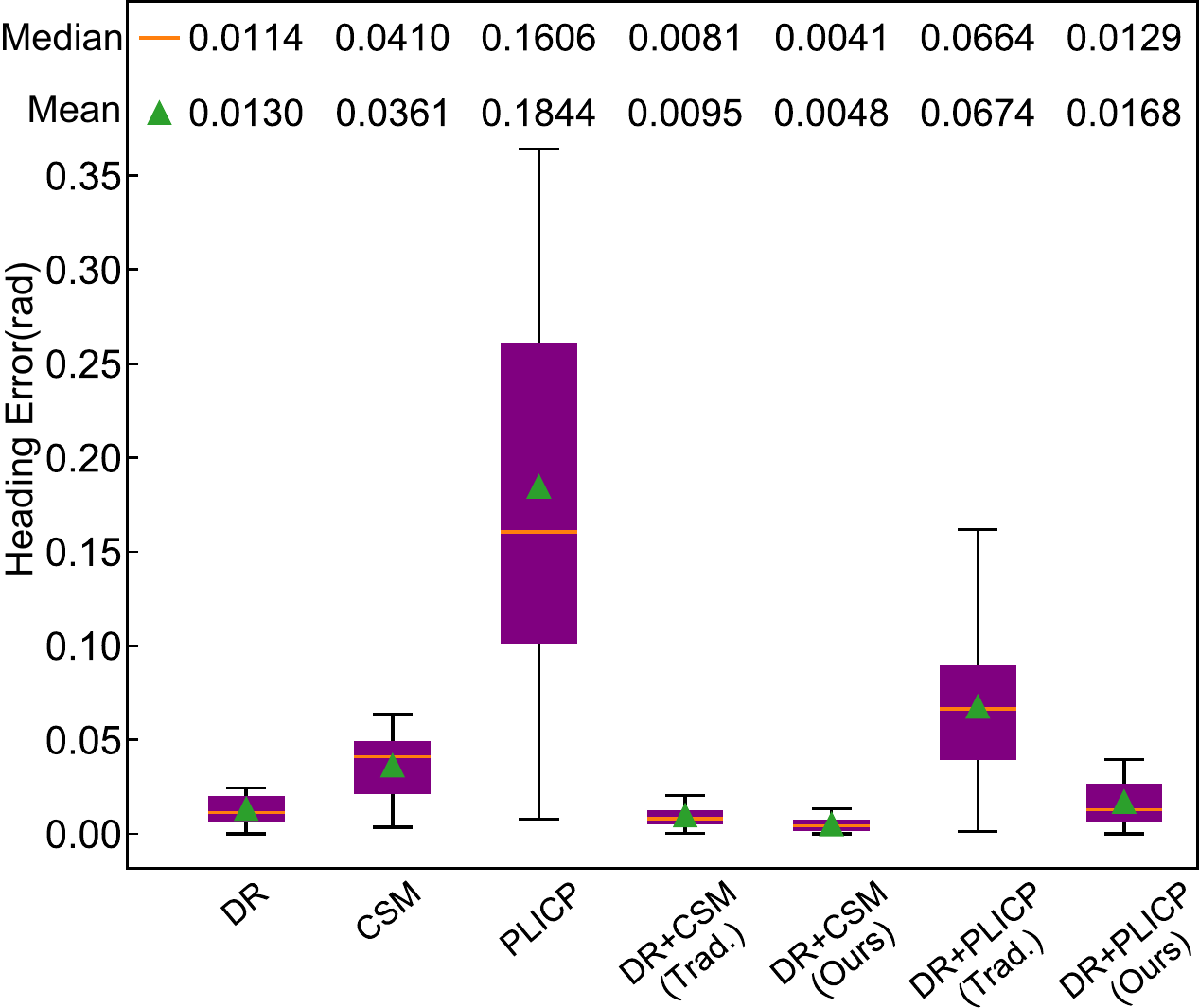}}
	{\caption{Localization accuracy using LiDAR odometry on real data in experienced environment.}}
	\label{fig:campus-experienced-acc}
\end{figure}
\begin{figure*}[!htb]
	\centering
	\begin{minipage}[c][4in]{4.9in}
		\subfloat[Trajectory comparison of different methods.]{%
			\label{fig:traj-unexp}
			\centering
			\includegraphics[width=4.9in]{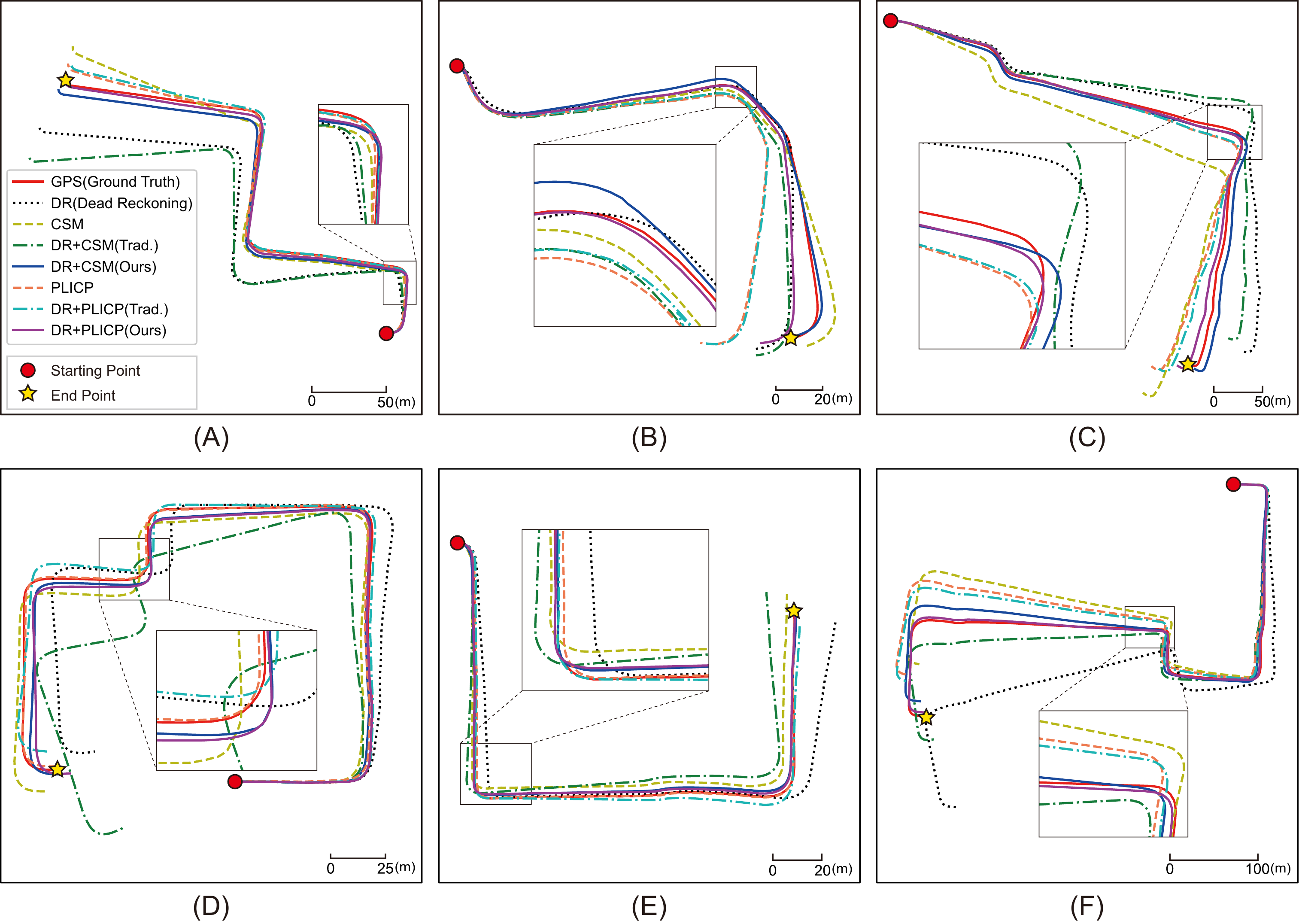}
			
		}
	\end{minipage}
	\hspace{\tabcolsep}
	\begin{minipage}[c][4in]{1.8in}
		\subfloat[Euclidean distance error.]{%
			\label{fig:accdis-unexp}
			\includegraphics[width=1.8in]{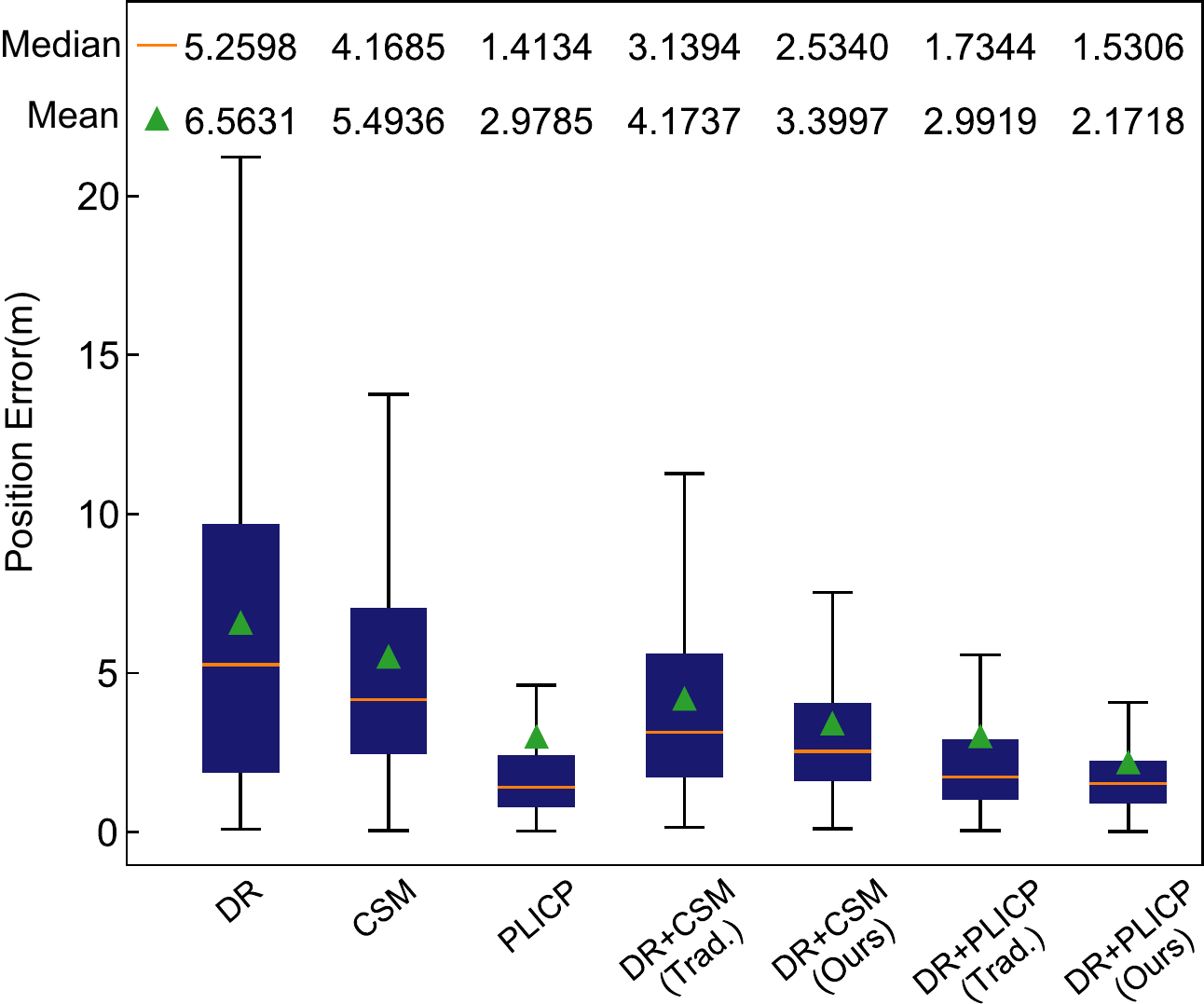}
		}
		
		\subfloat[Heading error.]{%
			\label{fig:accyaw-unexp}
			\includegraphics[width=1.8in]{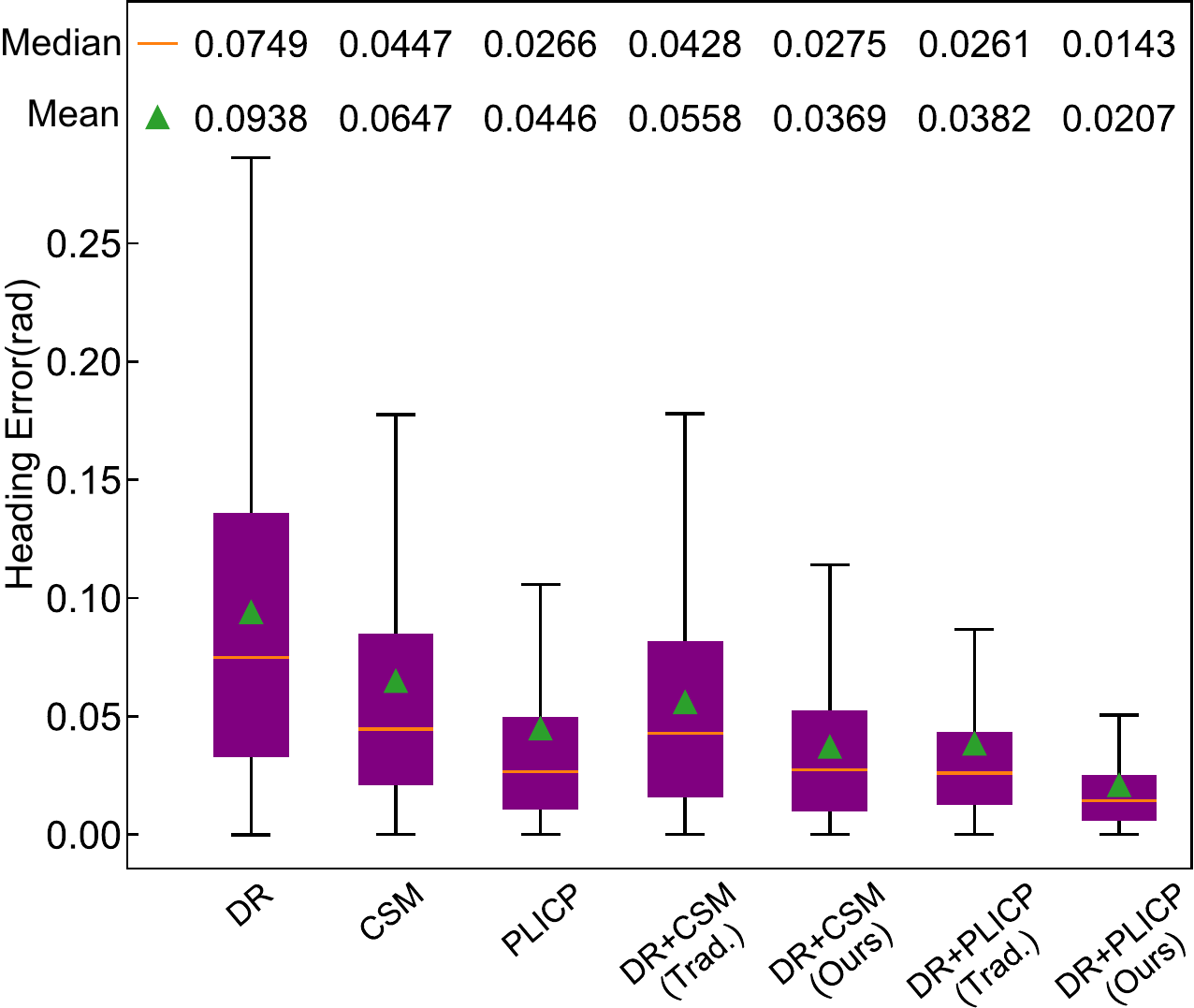}
		}
	\end{minipage}
	\caption{Localization accuracy using LiDAR odometry on real data in unexperienced environment.}
	\label{fig:campus-unexperienced-acc}
\end{figure*}
\subsubsection{Experimental Settings}
During the experiment, a new scan matching is triggered once the car moves ahead by 1.0 meter or the heading angle changes by 30 degrees since the last operation. In the training process, we use a constant step $T=100$, meaning that the program conducts forward prediction based on the current parameter set $\Theta$ for every $T$ steps. Then, a ground truth $x^g_T$ is obtained from the GPS/IMU suite and is used to adjust $\Theta$ through backpropagation along the sequence to minimize the error $J$ between $x^g_T$ and the predicted location $\mu^g_T$. However, the hyperparameter $\lambda$ of the loss function $J$ is set to 100.0 in this research to weight the errors in distance and angle.

\subsubsection{Learning Result of Scene-Aware Error Model}The CSM\cite{Olson2015} is used to examine our error model performance in different training stages.
During training, a new $\Theta$ is learned every $T$ steps with a ground truth location obtained. Such a procedure iterates until a limit condition is reached.
Below, we use ``epoch" to denote a single pass through the full training set, and let $\Theta_k$ to represent the learned parameter set at Epoch $k$.
At each specific scene, the predicted covariance error of LiDAR scan matching changes with $\Theta_k$.
This result is analyzed in Fig. \ref{fig:train-epoch-change-a}, where three scenes are selected, and the predicted covariance is represented by 2 standard deviation ovals and sampled scatter points.
With the initial parameter set $\Theta_0$, the predicted covariance of all three scenes shows quite similar shapes. As the number of epochs increases, the shapes vary differently, but they show a tendency of converging to their own stable states.
We use the parameter sets $\Theta_0$, $\Theta_2$, $\Theta_6$ and $\Theta_9$ to estimate the sequences of vehicle poses, which are drawn in Fig. \ref{fig:train-epoch-change-b} as trajectories A, B, C and D, respectively.
It is obvious that the localization error decreases progressively from trajectory A to D, demonstrating the efficiency of the learning procedure, where the accuracy of the predicted covariance error model is greatly improved.

\subsubsection{Localization Accuracy in Experienced Environments}
The localization accuracy of the proposed method is compared with dead reckoning, LiDAR odometry CSM\cite{Olson2009} and PLICP\cite{Censi2008}, and their conventional fusion-based method using covariance estimation in \cite{Olson2009} and \cite{Censi2007}. The sample trajectories estimated by these methods on testing data are shown in Fig. \ref{fig:traj-experienced}.
Compared with the traditional fusion method, the trajectories of our fusion method (solid line) are closer to the ground truth than the traditional methods (dashed dotted line).
With the GPS/IMU output as the ground truth, the position and heading error statistics of every 100-meter-long trajectory segment are plotted in Fig. \ref{fig:accdis-experienced} and Fig. \ref{fig:accyaw-experienced}.
For the fusion of DR and CSM, our method obtains a 12.7\% and 49.5\% reduction in the average Euclidean distance error and average yaw error, respectively; for the fusion of DR and PLICP, our method obtains a 48.1\% and 75.1\% reduction in the average Euclidean distance error and average yaw error, respectively.

\subsubsection{Localization Accuracy in Unexperienced Environments}

Similarly, localization trajectories on testing data of unexperienced regions are compared with other methods, as shown in Fig. \ref{fig:traj-unexp}.
The position and heading error statistics of every 100-meter-long trajectory segment are plotted in Fig. \ref{fig:accdis-unexp} and \ref{fig:accyaw-unexp}. From Fig. \ref{fig:traj-unexp}, we can see that the trajectories of our fusion methods denoted by solid lines are closer to the ground truth than the traditional trajectories drawn by dashed dotted lines. From a statistical perspective, our methods reduce the average Euclidean error by 18.5\%(DR+CSM) and 27.4\%(DR+PLICP) and the average yaw error by 33.9\%(DR+CSM) and 45.8\%(DR+PLICP).

To compare with the conventional method of error modeling, it is noteworthy that the measurement noise parameter $\sigma$ in Eq. \ref{eq:hessian} of the Hessian method and the likelihood $p$ in Eq. \ref{eq:sampling} of the sampling method has a strong effect on the covariance scale, which may lead to different fusion accuracies in RPF $\mathcal{F}$. Therefore, in the real data experiment section, we perform a grid search to rescale the covariance from the conventional method so that their best performance on the error scale can be used to compare with our method.

\section{Experiment on Visual Odometry}
\label{sec:VOExp}
\begin{figure}[!h]
	\centering
	\includegraphics[width=1.0\linewidth]{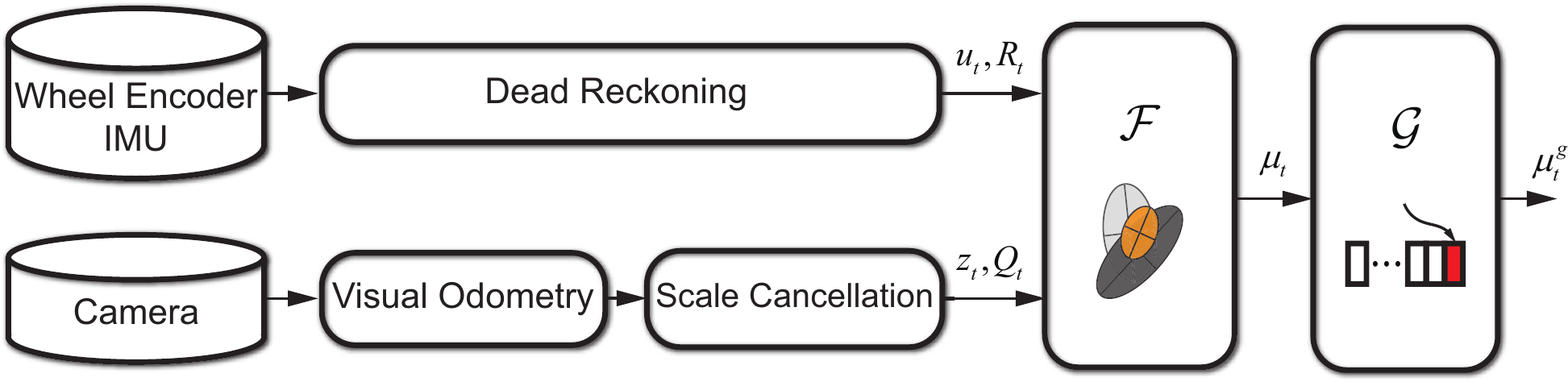}
	\caption{Processing flow of experiment on visual odometry.}
	\label{fig:processingflow-vo}
\end{figure}
\begin{figure*}[!h]
	\centering
	\begin{minipage}[c][4in]{4.8in}
		\subfloat[Trajectory comparison of different methods.]{%
			\centering
			\label{fig:votraj}
			\includegraphics[width=4.8in]{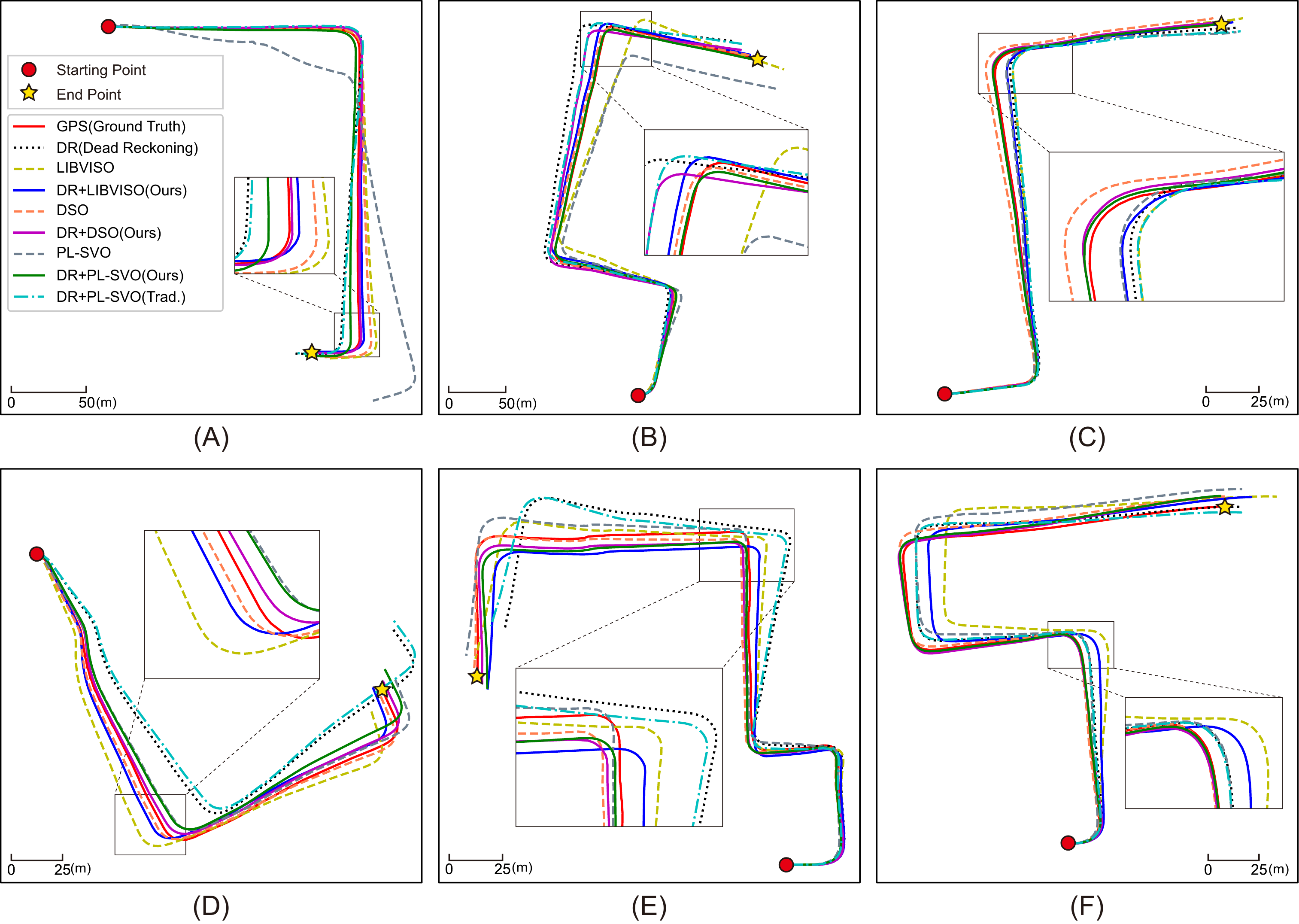}
		}
	\end{minipage}
	\hspace{\tabcolsep}
	\begin{minipage}[c][4in]{1.9in}
		\subfloat[Euclidean distance error.]{%
						\label{fig:voacc-dis}
			\includegraphics[width=1.9in]{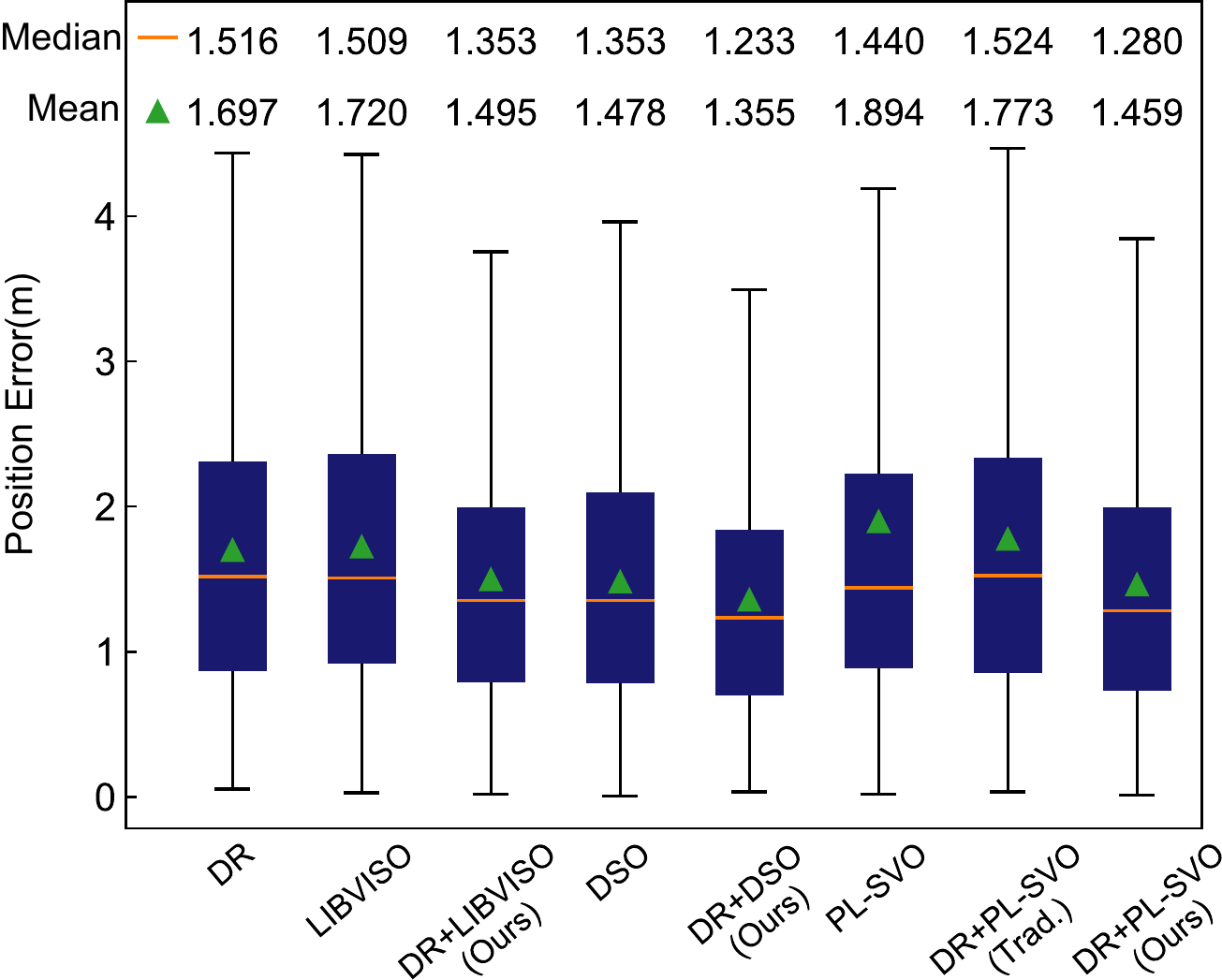}
		}
		
		\subfloat[Heading error.]{%
									\label{fig:voacc-yaw}
			\includegraphics[width=1.9in]{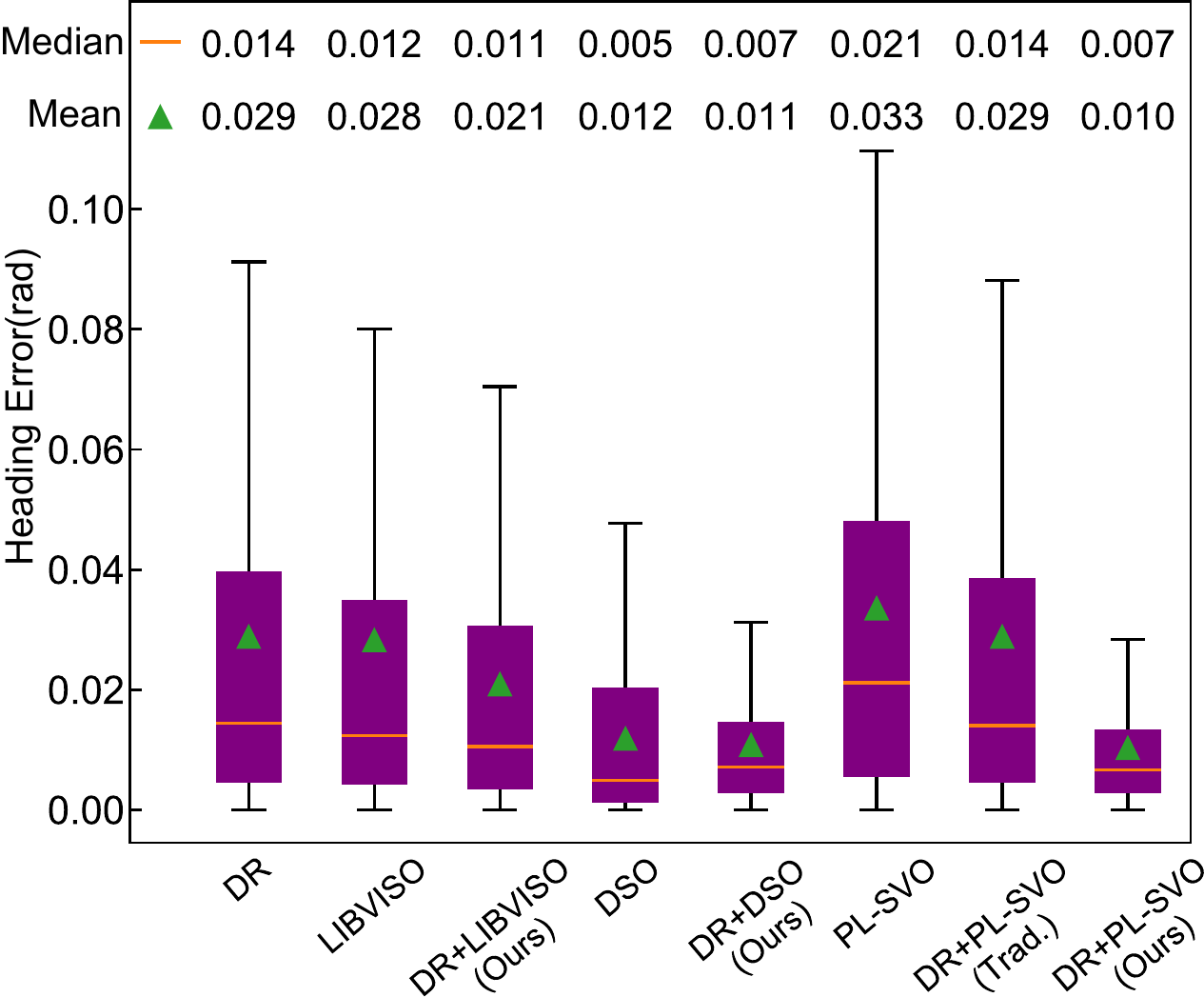}
		}
	\end{minipage}
		\label{fig:campus-unexperiencedVO-acc}
	\caption{Localization accuracy using visual odometry on real data in unexperienced environment.}

\end{figure*}
This is a supplementary experiment to prove that our method is also effective in odometry of other modalities except for LiDAR odometry.
An overview of the processing flow for fusing visual odometry is given in Fig. \ref{fig:processingflow-vo}. Similar to the experiment on LiDAR odometry, dead reckoning (DR) is used as the referred PSO module.
For the target visual odometry, three representative algorithms, LIBVISO\cite{Kitt2010}(feature-based method), DSO\cite{Engel2017}(direct method) and PL-SVO\cite{Gomez-ojeda2016}(a variant of SVO\cite{Forster2014}, semidirect method), are selected to perform error model learning. However, because there is no available error model for LIBVISO and DSO, the conventional error model comparison can only be performed on PL-SVO. Limited to sensor equipment, all of these visual odometries work in a monocular mode in our experiments.

\subsection{State Transition Model and Network Design}
Here, we use the same state definition as the experiments on LiDAR odometry, and the 6DoF results of visual odometry used in our experiments are projected to 3DoF in the same coordinate as GPS/IMU. However, as it is challenging for monocular visual odometries to output reliable scale information\cite{strasdat2010scale}, we need to customize our state transition model Eq. \ref{eq:statetransition} as follows:
\begin{equation}
\label{eq:statetransition-vo}
\left\{
\begin{aligned}
x_t &=u_t+\epsilon_t\qquad\quad\ (\epsilon_t \sim R_t) \\
z_t &=g(x_t)+\delta_t\qquad (\delta_t \sim Q_t) \\
\end{aligned}
\right.
\end{equation}
in which
\begin{equation}
g(x_t)=\frac{s(z_t)x_t}{s(x_t)}
\end{equation}
and $s(v)=\sqrt{v_x^2+v_y^2}$ is the function for calculating the translation scale of relative movement so that we can use an extended information filter to track this nonlinear state update. Steps 4-5 in Algorithm \ref{algo:LPF} should be modified as
\begin{equation}
\begin{aligned}
\Omega_t&=G_t^T\tilde Q_{t}G_t+\overline{\Omega}_t\\
\xi_t&=G_t^T\tilde Q_{t}(z_t-g(\overline\mu_t)+G_t\overline\mu_t)+\overline{\xi}_t
\end{aligned}
\end{equation}
where $G_t=g'(\overline\mu_t)$, $\overline\mu_t=u_t$.

For visual odometry, the real-time images can be used as scene information, so that a similar network architecture, as shown in Fig. \ref{fig:network-eng}, is used in this experiment. For the sake of training efficiency, we resize the grayscale image to the size of $115\times153$ (pixel), and the network structural parameters related to the input image size are also-modified accordingly.

\subsection{Real Data Experiment}
\subsubsection{Dataset}
This dataset was also collected using the platform, as shown in Fig. \ref{fig:jeepcar}. The following sensors are used: 1) a monocular camera is mounted above the windshield for visual odometry; 2) a wheel encoder and an IMU with lower precision are used for dead reckoning; and 3) a highly accurate GPS/IMU suite is used to obtain ground truth locations of the vehicle for model training and localization result evaluation.

Based on the good performance of the experiment on LiDAR odometry, we only challenge the experiments on visual odometry in an unexperienced environment to verify the extensibility of our method. Similarly, a large-scale dataset is collected in several different regions with a total mileage of approximately 10 km. Training data accounts for 60\% of the dataset, where there is almost no scene intersection with the remaining data for testing.
\subsubsection{Experimental Settings}
To compare trajectories from different methods synchronously, we keep the trigger behavior of DSO and align the trigger time of the other visual odometries LIBVISO and PL-SVO with DSO. In the training process, we use a constant step $T=100$, and the hyperparameter $\lambda$ of the loss function $J$ is set to 100.0 to balance the errors in distance and angle.
\subsubsection{Localization Accuracy}
The localization accuracy of the proposed method is compared with dead reckoning, DSO\cite{Engel2017}, LIBVISO\cite{Kitt2010}, PL-SVO\cite{Gomez-ojeda2016}, and the conventional fusion-based method using covariance estimation of PL-SVO in the author's open source code\footnote{https://github.com/rubengooj/pl-svo.git}. Their localization results are shown by several sample trajectories listed in Fig. \ref{fig:votraj}. We still use the accumulated error on every 100-meter-long trajectory segment in the test dataset to evaluate the localization accuracy of these methods, and the position and heading error statistics are plotted in Fig. \ref{fig:voacc-dis} and Fig. \ref{fig:voacc-yaw}.
In most cases, the fusion result trajectories of our method are closer to the GPS ground truth than the corresponding single-modal methods, as Fig. \ref{fig:votraj} shows, which indicates that the usage of our error model can explore their complementary information efficiently. For PL-SVO, of which we have access to the traditional error model, our method can obtain an error decrease of 17.7\% on the average Euclidean distance error, and 65.5\% on the average yaw error.

In contrast, the conventional method of PL-SVO error modeling achieves poor performance. In addition to the reasons analyzed in subsection \ref{subsec:Error Model}, another important reason lies in its derivation. PL-SVO optimizes the SE(3) pose using the left multiplicative perturbation model on its Lie group $\mathfrak{se}(3)$ so that its covariance on $\mathfrak{se}(3)$ needs to be mapped to SE(3) and SE(2) to be used in our state transition model as Eq. \ref{eq:statetransition-vo}. In this process, several nonlinear mappings need to be linearized, which makes this error model more inaccurate.

\section{Conclusion}
\label{conclusion}
In this research, a scene-aware error model is designed for LiDAR/visual odometry, and a localization fusion framework is developed to fuse the results using such an error model. Moreover, an end-to-end learning method is devised to train the error model in the proposed localization fusion framework.

We thoroughly evaluate the proposed method on simulation data to verify its adaptability at various simple but typical scenes and on real data to examine its efficiency in real-world situations. The experimental results demonstrate that the proposed method is efficient in learning the CNN-based error model, and the localization accuracy based on such models is superior compared with the fusion accuracy of the other traditional methods.

Future work will focus on the following limitations of the proposed method.

1) Gradient vanishing problem. This is a general problem of training RNNs (recurrent neural networks)\cite{mikolov2010recurrent}\cite{pascanu2013difficulty}. Apparently, the error model learning process of our method is similar to the typical RNN. When training using a long trajectory with a large number of iterations, we expect the error information to be amplified by continuous rotation, whereas sometimes it is also likely to be overwhelmed.

2) Optimization of the training trigger strategy. In our experiment, every backpropagation is performed after forward prediction with fixed time steps, which is convenient for off-line batch training. However, such a method cannot fit Eq.\ref{gtcondition} properly and makes it inefficient to be extended to on-line learning.

3) Hyperparameter setting. The hyperparameter $\lambda$ in the loss function (Eq. \ref{lossfunc}) is another important factor for training performance.
Manually $\lambda$ in the experiment above is selected after many attempts. This troublesome but necessary procedure must be performed for different datasets.

\appendices

\setcounter{equation}{0}
\setcounter{algorithm}{0}

\renewcommand{\thealgorithm}{\thesection-\arabic{algorithm}}
\renewcommand{\theequation}{\thesection-\arabic{equation}}

\section{The Original Information Filter}

\label{appendix-i}

Assume that the state transition and measurement probabilities are governed by the following linear Gaussian equations:
	\begin{equation}
   	\label{orginformationfilter}
	\begin{aligned}
	x_t&=A_tx_{t-1}+B_tu_t+\epsilon_t\\
	z_t&=C_tx_t+\delta_t
	\end{aligned}
	\end{equation}
where $u_t$ and $z_t$ are the control and measurement at time $t$,
$\epsilon_t$ and $\delta_t$ denote their Gaussian noise with covariances $R_t$ and $Q_t$, respectively.

Probabilistic estimation of $x_t$ using a Gaussian filter finds a mean pose $\mu_t$ and a covariance matrix $\Sigma_t$.
Whereas using an information filter \cite{Thrun2005}, the Gaussian distribution is represented in canonical representation, and the problem is to estimate an information vector $\xi_t=\Sigma_t^{-1}\mu_t$ and an information matrix $\Omega_t = \Sigma_t^{-1}$, which is described in Algorithm \ref{informationfilter}.

   	\begin{algorithm}[h]
   		\caption{Information Filter($\xi_{t-1},\Omega_{t-1},u_t,z_t$)}
   		\label{informationfilter}
   		\begin{algorithmic}[1]
   			\State $\overline{\Omega_t}=(A_t\Omega_{t-1}^{-1}A_t^{T}+R_t)^{-1}$
   			\State $\overline{\xi_t}= \overline{\Omega_t}(A_t\Omega_{t-1}\xi_{t-1}+B_tu_t)$
   			\State $\Omega_t=C_t^TQ_t^{-1}C_t+\hat{\Omega_t}$
   			\State $\xi_t=C_t^TQ_t^{-1}z_t+\overline{\xi_t}$
   			\State \Return $\xi_t,\Omega_t$
   		\end{algorithmic}
   	\end{algorithm}

\section{Derivation of the RPF Algorithm}
In this research, we defined the state transition and measurement function in Eq. \ref{eq:statetransition}. Compared with the definitions in Eq. \ref{orginformationfilter}, we have $A_t=\mathbf{0}$, $B_t=I$, $C_t=I$, where $I$ is the identical matrix. Substituting these matrices in Algorithm \ref{informationfilter}, we obtain lines 2-5 of Algorithm \ref{if}.

\bibliographystyle{IEEEtran}
\bibliography{bibtex/bib/ref.bib,bibtex/bib/multimodal-localization.bib,bibtex/bib/pbo-LO.bib,bibtex/bib/pbo-VO.bib,bibtex/bib/pbo-others.bib,bibtex/bib/errormodel.bib}

\end{document}